\pdfoutput=1

\documentclass[11pt]{article}

\usepackage[preprint]{coling}

\usepackage{times}
\usepackage{latexsym}

\usepackage[T1]{fontenc}

\usepackage[utf8]{inputenc}

\usepackage{microtype}

\usepackage{inconsolata}

\usepackage{graphicx}

\usepackage{wasysym}
\usepackage{array}
\newcolumntype{H}{>{\setbox0=\hbox\bgroup}c<{\egroup}@{}}
\title{Explanation Regularisation through the Lens of Attributions}

\author{Pedro Ferreira~\textsuperscript{1} \quad Ivan Titov~\textsuperscript{1,2} \quad Wilker Aziz~\textsuperscript{1} \\
$^1$~University of Amsterdam \quad $^2$~University of Edinburgh \\
\texttt{ \{ p.m.ferreira, w.aziz \}@uva.nl} \quad \texttt{ititov@inf.ed.ac.uk}
}

\begin{document}
\maketitle

\begin{abstract}

Explanation regularisation (ER) has been introduced as a way to guide text classifiers to form their predictions relying  on input tokens that humans consider plausible.
This is achieved by introducing an auxiliary explanation loss that measures how well the output of an input attribution technique for the model agrees with human-annotated rationales.  
The guidance appears to benefit performance in out-of-domain (OOD) settings, presumably due to an increased reliance on `plausible' tokens. 
However, previous work has under-explored the impact of guidance on that reliance, particularly when  reliance is measured using attribution techniques different from those used to guide the model.  
In this work, we seek to close this gap, and also explore the relationship between  reliance on plausible features and OOD performance. 
We find that the connection between ER and the ability of a classifier to rely on plausible features has been overstated and that a stronger reliance on plausible tokens does not seem to be the cause for  OOD improvements.
\footnote{Source code available at \url{https://github.com/PedroMLF/ER_through_the_lens_of_attributions}.}\looseness=-1

\end{abstract}
\section{Introduction}

\begin{figure}[t]
    \centering
    \includegraphics[width=1.\linewidth]{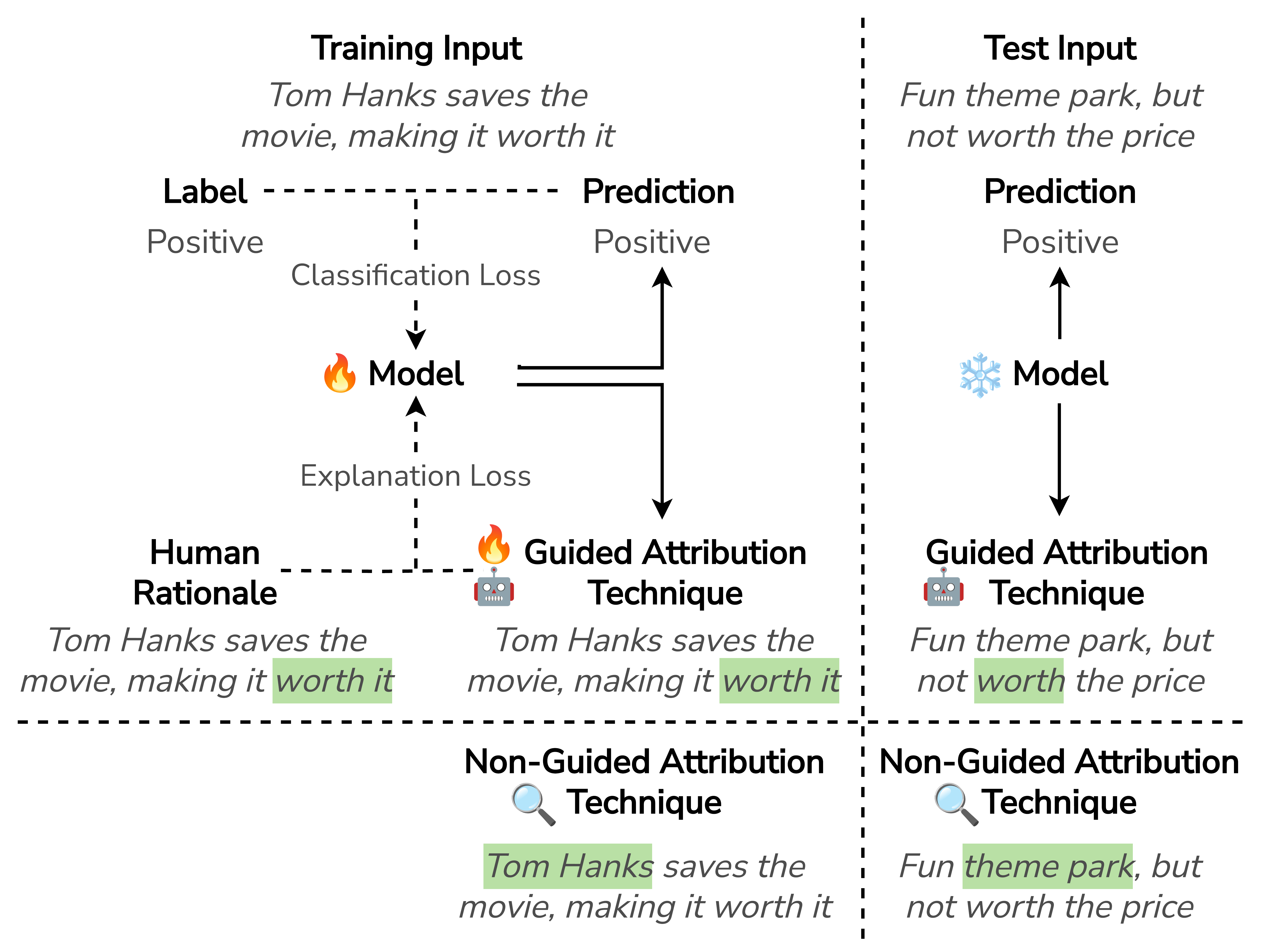}
    \caption{ER (top-left)  minimises a \emph{classification} and an \emph{explanation} loss. The latter uses an input attribution technique to obtain a machine rationale for the prediction, and penalises differences between that and the human rationale. Guiding the model to rely in its predictions on plausible tokens is expected to help the classifier at test time (top-right), when no human rationale is available.
    Bottom: even though the attribution technique used for guidance shows human-like rationales, the model may in fact rely on different (non-plausible) tokens, as other attribution techniques might reveal.}
    \label{fig:er_summary}
\end{figure}

In explanation regularisation \citep[ER;][\ia]{ross2017rrr, ghaeini2019saliency, liu2019incorporating, ismail2021improving, joshi-etal-2022-er, pruthi2022evaluating, stacey2022supervising}, a text classifier is encouraged to inform its predictions by tokens included in a human rationale for the  label.
This is achieved with an auxiliary `explanation loss' that penalises differences between the output of a \emph{guided attribution technique} (\eg, a relevance map based on top-layer attention) and the annotated \emph{human rationale} (see Figure \ref{fig:er_summary}).
Compared to their counterparts trained without rationales, these ER models have been reported to improve classification, including for out-of-domain (OOD) inputs \citep{joshi-etal-2022-er,stacey2022supervising,madani2023refer}, with this improved robustness being ascribed to increased reliance of ER models on \emph{plausible tokens} (\ie, those in the human rationale for the label).

While ER is assumed to encourage classifiers to rely more on plausible tokens, this has not been carefully verified in previous work. 
In particular, prior analyses have focused on guided attributions---\ie, those \emph{explicitly supervised} to resemble human rationales through the explanation loss---finding them to be better aligned with human rationales \citep[\ia]{mathew2021hatexplain,stacey2022supervising,madani2023refer}.
Although impact on guided attributions is a necessary condition, this impact is not sufficient to demonstrate that the classifier  relies more on plausible tokens.  
We argue that, to confirm  an increased reliance, it is also necessary to examine the impact of guidance on attributions obtained through other \emph{non-guided} attribution techniques (see Figure \ref{fig:er_summary}).
If these  are unaffected, it would suggest that the model is `hacking' \citep{skalse2022defining} the explanation loss rather than genuinely increasing its reliance on plausible tokens.\looseness=-1

In this work, we analyse \emph{(i)} ER's ability to make classifiers rely on plausible tokens; and  \emph{(ii)} the relationship between  plausibility and robustness to OOD conditions.
We start by studying ER with jointly optimised losses (\S\ref{subsec:joint_optimisation}).
We find that -- unlike the guided attribution techniques -- there is little to no evidence that attributions 
by any other non-guided attributions are affected by ER, suggesting the loss `hacking'.  
In fact, we find that meaningful impact on input attributions is only achievable by ensuring a low explanation loss, for example via constrained optimisation (see \S \ref{subsec:constrained_optimisation}).
However, this is effective only for `global' guided attribution techniques (which capture the information `flow' across all or most components of the model), and, crucially, at the expense of classification performance.\looseness=-1

\paragraph{Contributions.}
We show that the connection between ER and the ability of a classifier to rely on plausible tokens has been overstated. In particular, we find that the choice of guided attribution technique is critical for this assumption: 
local attribution techniques, such as top-layer attention, can be `hacked', minimising the explanation loss without much evidence of increased reliance on  plausible tokens. 
Moreover, we find that OOD classification performance degrades for ER models that (over-)rely on plausible tokens, hinting at the need of finding better design choices for ER.

\section{Related Work}
\label{sec:related_work}

\paragraph{Learning from explanations.}
Human explanations (\eg, token-level rationales, as in Figure \ref{fig:er_summary}) have been used to improve text classifiers \citep{hartmann2022survey,hase-bansal-2022-models}.
Examples include \emph{select-predict} `pipelines', where an explanation output by a first module serves as input to a classifier \citep{camburu2018snli}, and \emph{multi-task} settings, where a classifier and a rationale extractor  are jointly trained \citep{carton2022learn,chan2022unirex}.
~
In contrast, we are interested in \emph{explanation regularisation}, where, rather than having additional model components trained to perform rationale extraction, a model is trained to align \emph{attributions} provided by a differentiable attribution technique with human rationales \citep[\ia]{ross2017rrr, ghaeini2019saliency, liu2019incorporating, rieger2020interpretations, mathew2021hatexplain,pruthi2022evaluating, stacey2022supervising}.
Most work on ER focuses on in-domain data, however, there is evidence that ER can improve robustness to OOD data, for  text \cite{rieger2020interpretations, joshi-etal-2022-er, stacey2022supervising}, and image classification \cite{chefer2022robustness,rao2023studying}. 
Other works identify difficulties in  using explanations to learn better classifiers and propose methods to improve the compromise between  classification performance and model explainability. These works differ from ours by: \emph{(i)} not using human-annotated explanations \citep{plumb2020regularizing}; or \emph{(ii)} by departing significantly from the ER formulation, for example, by using a multi-task setup with a separate explainer \citep{carton2022learn} or by not employing a trainable explainer and instead incorporating human explanations through contrastive learning \citep{resck2024exploring}.
We, instead, focus on ER and explore its impact on the plausibility of input attributions and how that, in turn, impacts classification performance both in- and out-of-domain.

\paragraph{Attribution techniques.} 

In \emph{vector-based} techniques, model components are directly used to obtain input attributions. The simplest approach is to use attention weights \citep[\ia]{clark-etal-2019-bert}, despite conflicting evidence on its usefulness for this purpose \citep{jain2019attention,serrano2019attention,wiegreffe2019attention,pruthi2020learning}.
Later works \citep{kobayashi-etal-2020-attention, kobayashi-etal-2021-incorporating, ferrando2022measuring, modarressi2022globenc} incorporate information about the magnitude of input vectors and the influence of other components of the Transformer layer \citep{vaswani2017attention}.
These are examples of \emph{local} techniques, where attributions are based on the dynamics of a single layer.\looseness=-1

It is also possible to obtain a \emph{global} analysis of the model, by including the dynamics of its multiple layers.
One example is \emph{attention-rollout} \citep{abnar2020quantifying}, which recursively aggregates attention weights across layers.
Other examples of global vector-based techniques are GlobEnc \citep{modarressi2022globenc}, ALTI \citep{ferrando2022measuring}, and DecompX \citep{modarressi2023decompx}.
Global attributions can also be obtained from \emph{gradient-based} approaches \citep[\ia]{kindermans2016investigating,shrikumar2017learning,sundararajan2017axiomatic}, or a mix of attention and gradients \citep{chefer2021transformer,qiang2022attcat}.\looseness=-1

\section{Explanation Regularisation}
\label{sec:er}

In ER \citep[\ia]{joshi-etal-2022-er, stacey2022supervising, pruthi2022evaluating}, we assume the availability of training data $\mathcal{D} = \{(x_i, y_i, e_i)_{i=1}^N \}$, for each data point, $x_i$ is the input text, $y_i$ is the target label, and $e_i$ is a human rationale (specifically, a subset of tokens of the input regarded as a justification for the label).
Human rationales can be obtained from annotators \citep{socher-etal-2013-recursive, camburu2018snli, rajani-etal-2019-explain, mathew2021hatexplain} or through heuristics  \citep{liu2019incorporating,rieger2020interpretations}, and are only required during training.

An ER classifier minimises a joint loss
\begin{equation}
    \mathcal{L}_{\mathrm{cls}}(\theta) + \lambda ~\mathcal{L}_{\mathrm{expl}}(\theta),
    \label{eq:er_joint}
\end{equation}
where $\mathcal{L}_{\mathrm{cls}}$ is the regular classification loss and $\mathcal{L}_{\mathrm{expl}}$ is the \emph{explanation loss}, with a hyperparameter $\lambda \in \mathbb R_{\ge 0}$ controlling the contribution of the latter.
The expectation is that $\mathcal{L}_{\mathrm{expl}}$ leads the model to 
rely in its decisions on more plausible tokens.

The first term is the cross-entropy loss $\mathcal{L}_{\mathrm{cls}}(\theta) = \mathbb E_{(x, y)\sim \mathcal D}[ \mathcal{L}_{\mathrm{ce}} (C_\theta (x), y) ]$, 
where $C_\theta$ is a trainable classifier.
The second term, $\mathcal{L}_{\mathrm{expl}}(\theta) = \mathbb E_{(x, y, e)\sim \mathcal D}[ \mathcal{S}(E (C_\theta, x), e) ]$, employs an attribution technique in order to obtain a relevance map  $E(C_\theta, x)$ for the classifier's output and penalises the classifier by this map's deviation from the human rationale, as assessed by a  function $\mathcal S$ such as mean absolute error or KL divergence \citep{joshi-etal-2022-er}.
~
We  refer to $E$ as \emph{the guided attribution technique}.
Besides being differentiable, $E$  is required to be memory- and time-efficient.
Examples of techniques used in ER include gradient-based \citep{ross2017rrr, ghaeini2019saliency, chefer2022robustness}, vector-based  \citep{mathew2021hatexplain, pruthi2022evaluating, stacey2022supervising, fernandes2022scaffold}, and perturbation-based approaches \citep{ying2022visfis}. We follow \citet{joshi-etal-2022-er} and \citet{stacey2022supervising} and use top-layer attention  and \textsc{InputXGradient} \citep{shrikumar2017learning}. In addition, we also experiment with attention-rollout \citep{abnar2020quantifying}.

\section{Experimental Setting}
\label{sec:experimental_settings}

\paragraph{Data.}
We use SST-2 \citep{socher-etal-2013-recursive} as training  and in-domain data, following the heuristic proposed by \citet{carton2020evaluating} to obtain instance-level rationales. For OOD data with rationales we use Movies \citep{zaidan-eisner-2008-modeling, deyoung2020eraser}, and annotate Yelp-50, a subset of Yelp, as described in Appendix \ref{sec_appendix:extra_ood_annotations}. For OOD without rationales annotations we use Amazon Reviews (`Movies and TV' split) \citep{hou2024bridging}, IMBD \citep{maas2011learningvectors}, and Yelp \citep{zhang2015character}.
For more details on data refer to Appendix \ref{subsec_appendix:sst_data}.

\paragraph{Model.}
All experiments use HuggingFace Transformers' \citep{wolf-etal-2020-transformers} \textsc{BigBird-roberta-base} model as the pre-trained contextual embedding encoder model \citep{zaheer2020big}. This choice follows previous works \citep{chan2022unirex, joshi-etal-2022-er, madani2023refer}. For more training details refer to Appendix \ref{sec_appendix:training}.

\paragraph{Scoring Function.}
We follow \citet{joshi-etal-2022-er} and use mean absolute error for $\mathcal{S}$ in $\mathcal{L}_{\mathrm{expl}}(\theta)$.

\paragraph{Target Annotations.}
All datasets with rationales are annotated at the instance level with binary vectors, whose  dimensionality equals the number of tokens. A value of 1 indicates relevant, highlighted tokens.
In order to obtain a target for the scoring function $\mathcal{S}$, the instance-level vector is normalized to sum to one by dividing each element by the total number of highlighted tokens in the instance when using top-layer attention and attention-rollout \citep{joshi-etal-2022-er, stacey2022supervising}. For \textsc{InputXGradient} we use the original binary values.\looseness=-1

\paragraph{Attribution Techniques.} We make use of one local technique, namely, top-layer attention (\textsc{Att}), and 5 global techniques, namely, attention rollout \citep[\textsc{AttR};][]{abnar2020quantifying}, \textsc{InputXGradient} \citep[\textsc{IxG};][]{shrikumar2017learning}, \textsc{ALTI} \cite{ferrando2022measuring}, and \textsc{DecompX} \cite{modarressi2023decompx} with (\textsc{DX-C}) and without (\textsc{DX}) the classification head.
\footnote{For \textsc{Att} and \textsc{AttR}, we average attention weights across heads. For \textsc{IxG} we use the Captum package \citep{kokhlikyan2020captum}. For \textsc{DecompX} and \textsc{ALTI} we adapt the original code.\looseness=-1}
For guidance we use \textsc{Att}, \textsc{AttR} or \textsc{IxG}, as those are  memory- and time-efficient enough to be used during training. The remainder are used only for analysis.
All techniques are further described in Appendix \ref{sec_appendix:attribution_techniques}.

\paragraph{Plausibility Metrics.}
To assess how well input attributions reproduce the human annotated rationales, we use three metrics introduced in \citet{fomicheva2021eval4nlp}, and described in Appendix \ref{sec_appendix:plausibility_metrics}: AUC Score, Average Precision; and Recall@k.\looseness=-1

\paragraph{Constrained Optimisation.} 
To study classifiers whose guided attributions are as plausible as they can be (through the lens of the explanation loss), we can reimagine ER as a constrained optimisation problem: 
minimise classification loss subject to a bound $b$ on the explanation loss,
\begin{equation}\label{eq:lag}
        \begin{aligned}
        \min_\theta & \quad  \mathcal{L}_{\mathrm{cls}} (\theta) &\mathrm{s.t.} & \quad \mathcal{L}_{\mathrm{expl}}(\theta) \le b \;,
        \end{aligned}
\end{equation}
with $b$ set to a value close to $\min_{\theta}~\mathcal{L}_{\mathrm{expl}}(\theta)$ -- a minimiser of the explanation loss, taken in isolation. We approach this via Lagrangian relaxation \citep{boyd2004convex}, with implementation choices detailed in Appendix \ref{sec_appendix:training}.

\section{Results}
\label{sec:results}

\begin{table}[h!]
    \centering
    \small
    \begin{tabular}{l l c cc}
    \toprule
     ER & Attr. & F1 & Guided & Non-Guided \\ 
     \midrule
   \multirow{2}{*}{\emph{Joint}  (\S\ref{subsec:joint_optimisation})} & Local  & \checkmark & \checkmark  & $\varnothing$ \\
      & Global & \checkmark & $\varnothing$ & $\varnothing$ \\
     \midrule
     \multirow{2}{*}{ \emph{Constr} (\S\ref{subsec:constrained_optimisation})} & Local  & \checkmark &  \checkmark   & $\varnothing$ \\
      & Global &  \newcrossmark & \checkmark    & \checkmark    \\
     \bottomrule           
\end{tabular}
    \caption{Summary of the effect ($\checkmark$  positive, $\newcrossmark$ negative, or $\varnothing$ nil) of local or global guidance, across Joint and Constrained setups, in terms of OOD classification  (F1) and impact on guided and non-guided attributions.}
    \label{tab:results_summary}
\end{table}

Table \ref{tab:results_summary} presents a brief overview of our observations, which are discussed in detail in the following subsections.   In \S\ref{subsec:joint_optimisation}, we find that despite positively impacting the OOD generalisation of the classification task, joint ER has limited impact on input attributions. In fact, only local guidance is able to better predict plausible rationales, and that effect is only observed for the guided technique itself. 
With global guidance, we find that the explanation loss is under-optimised. To understand what is happening, in \S\ref{subsec:constrained_optimisation} we reinterpret ER as constrained optimisation, where the explanation loss is constrained to be low.
At this point, we find it to be possible to affect input attributions meaningfully, but only when guiding a global attribution technique and, crucially, at the expense of classification performance.
In a nutshell, local guidance is being `hacked': attributions are modified locally as to optimise the explanation loss without affecting the features used for classification (intuitively, the model `hides' non-plausible computations where the local attribution technique cannot `see', such as in lower layers). Global guidance, on the other hand, seems much harder to `hack': optimising it well requires modifying computations performed in different layers across the model, more strongly restricting the features used for classification to rationale tokens. This, however, tends to worsen classification performance. 
Finally, we find in \S\ref{subsec:predict_ood_performance} that the disconnect between (over-)reliance on plausible tokens and OOD classification performance prevents us from systematically using features available for model selection to find models that perform best OOD.

\subsection{Joint Optimisation}
\label{subsec:joint_optimisation}

We start by studying the most common ER setup (\S\ref{sec:er}), where the two losses are jointly optimised.

\begin{figure*}
    \centering
    \includegraphics[width=1.0\linewidth]{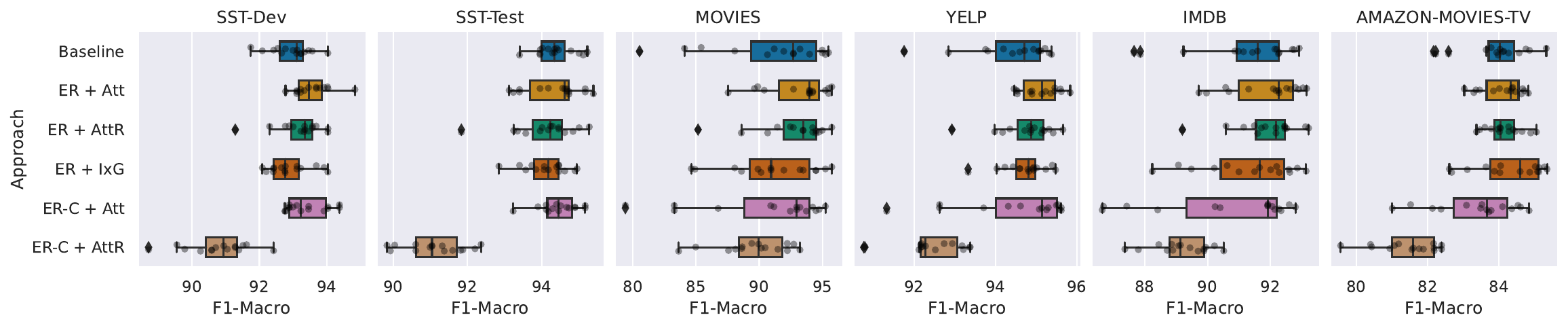}
    \caption{F1-Macro scores ($\uparrow$). \textsc{ER + Att} uses attention as the guided attribution technique, \textsc{ER + AttR} uses attention-rollout, and \textsc{ER + IxG} uses \textsc{InputXGradient}. The \textsc{C} prefix indicates a constrained model. Results correspond to 15 seeds. \textsc{ER-C + IxG} is not shown for improved clarity and can be seen in Appendix Figure \ref{fig:sa_results_constrained}.}
    \label{fig:sa_results}
\end{figure*}

\paragraph{Joint ER improves OOD classification.}
As observed in Table \ref{tab:sa_results}, joint ER improves OOD average classification performance.
This is more noticeable for guidance with local attention (\textsc{ER+Att}), with guidance that uses global attribution techniques (\textsc{ER+AttR} and \textsc{ER+IxG}) showing smaller improvements.
If we consider variance across runs, and visualise the distribution of results across seeds (Fig. \ref{fig:sa_results}), it is noticeable how the effect on the  average improvement is exacerbated by  runs that perform poorly.
In fact, this seems to indicate that one of the merits of ER is to converge less often to models that generalise poorly to OOD conditions.

\begin{table*}
\centering
\small
\begin{tabular}{l lcccccc}
\toprule
 & & \textit{SST Dev} & \textit{SST Test} & \textbf{Movies} & \textbf{Yelp} & \textbf{IMDB} & \textbf{Amazon-M-TV} \\
\midrule
& \textsc{Baseline} & 92.95 $\pm$ 0.57 & 94.36 $\pm$ 0.48 & 91.01 $\pm$ 4.38 & 94.41 $\pm$ 0.99 & 91.20 $\pm$ 1.63 & 83.89 $\pm$ 0.89 \\
\midrule
\multirow{3}{*}{\shortstack{J \\ \S\ref{subsec:joint_optimisation}}} & \textsc{Attention} & 93.56 $\pm$ 0.50 & 94.33 $\pm$ 0.72 & 93.10 $\pm$ 2.44 & 95.09 $\pm$ 0.44 & 91.87 $\pm$ 1.10 & 84.10 $\pm$ 0.56 \\
& \textsc{+ Rollout} & 93.19 $\pm$ 0.68 & 94.09 $\pm$ 0.82 & 92.72 $\pm$ 2.71 & 94.77 $\pm$ 0.66 & 91.88 $\pm$ 0.99 & 84.15 $\pm$ 0.47 \\
& \textsc{IxG} & 92.91 $\pm$ 0.58 & 94.13 $\pm$ 0.56 & 91.11 $\pm$ 3.37 & 94.72 $\pm$ 0.54 & 91.30 $\pm$ 1.44 & 84.29 $\pm$ 0.92 \\
\midrule
\multirow{3}{*}{\shortstack{C \\ \S\ref{subsec:constrained_optimisation}}} & \textsc{Attention} & 93.41 $\pm$ 0.57 & 94.45 $\pm$ 0.49 & 90.59 $\pm$ 4.80 & 94.52 $\pm$ 1.30 & 90.65 $\pm$ 2.17 & 83.38 $\pm$ 1.11 \\
& \textsc{+ Rollout} & 90.78 $\pm$ 0.90 & 91.11 $\pm$ 0.77 & 89.56 $\pm$ 2.67 & 92.40 $\pm$ 0.78 & 89.19 $\pm$ 0.85 & 81.44 $\pm$ 0.81 \\
& \textsc{IxG} & 90.64 $\pm$ 0.90 & 91.59 $\pm$ 1.46 & 75.80 $\pm$ 15.1 & 88.99 $\pm$ 6.17 & 83.35 $\pm$ 7.58 & 79.95 $\pm$ 2.57 \\
\bottomrule
\end{tabular}
\caption{F1-Macro ($\uparrow$) and standard deviation for \textit{in-domain (SST)} and \textbf{OOD data}. \textsc{Attention} corresponds to an ER (J)oint or (C)onstrained model that uses attention as the guided attribution technique, \textsc{+Rollout} to a model that uses attention-rollout, and \textsc{IxG} to a model that uses \textsc{InputXGradient}. Results are averages of 15 seeds.}
\label{tab:sa_results}
\end{table*}

\paragraph{Only local guidance improves rationale extraction performance.}
In addition to classification, ER models are trained to align guided attributions to human rationales. Hence, compared to the baseline, we expect an impact on the plausibility of the corresponding guided attributions.
This effect can be observed by comparing the \colorbox[HTML]{C0C0C0}{\emph{highlighted}} cells in Table \ref{tab:sa_plausibility_auc_id} with the corresponding baseline.
Our expectation is met \emph{only} with local guidance, with  average AUC plausibility increasing from 35.6 (baseline) to 78.6 (w/ \textsc{Att}) in-domain and similar, but more modest, improvements OOD (Table \ref{tab:sa_plausibility_auc_ood}).

\begin{table}[t]
    \centering
    \small
    \begin{tabular}{lcc @{\hskip 2em} cccc}
        \toprule
        & \multicolumn{6}{c}{\textbf{SST-Dev}} \\ \cmidrule{2-7}
        & \rot{90}{\textsc{Att}} & \rot{90}{\textsc{AttR}} & \rot{90}{\textsc{IxG}} & \rot{90}{\textsc{ALTI}} & \rot{90}{\textsc{DX-C}} & \rot{90}{\textsc{DX}} \\
        \midrule
        \textsc{Baseline} & 35.6 & 27.9 & 50.7 & 59.0 & 58.5 & 57.5 \\
        \midrule
        \multicolumn{7}{c}{\emph{Joint ER} (\S\ref{subsec:joint_optimisation})} \\
        \midrule
        \textsc{w/ Att} & \cellcolor[HTML]{C0C0C0}\emph{78.6} & 27.9 & 50.6 & 64.5 & 58.4 & 57.8 \\
        \textsc{w/ AttR} & 37.0 & \cellcolor[HTML]{C0C0C0}\emph{28.7} & 50.8 & 60.2 & 58.6 & 57.4 \\
        \textsc{w/ IxG} & 44.6 & 27.6 & \cellcolor[HTML]{C0C0C0}\emph{53.8} & 61.7 & 58.0 & 57.6 \\
        \midrule
        \multicolumn{7}{c}{\emph{Constrained ER} (\S\ref{subsec:constrained_optimisation})} \\
        \midrule
        \textsc{w/ Att} & \cellcolor[HTML]{C0C0C0}\emph{87.6} & 28.2 & 51.5 & 68.3 & 57.6 & 56.8 \\
        \textsc{w/ AttR} & 63.2 & \cellcolor[HTML]{C0C0C0}\emph{89.1} & 67.6 & 87.9 & 76.9 & 81.3 \\
        \textsc{w/ IxG} & 63.9 & 42.9 & \cellcolor[HTML]{C0C0C0}\emph{81.5} & 74.8 & 61.3 & 64.8 \\
        \midrule
        \multicolumn{7}{c}{\emph{$\mathcal{L} = \mathcal{L}_{\mathrm{expl}}$} (\S\ref{subsec:joint_optimisation})} \\
        \midrule
        $\mathcal L_{\text{expl}}$ (A) & \cellcolor[HTML]{C0C0C0}\emph{89.0} & 29.6 & 53.2 & 78.2 & 52.0 & 53.0 \\
        $\mathcal L_{\text{expl}}$ (R) & 81.1 & \cellcolor[HTML]{C0C0C0}\emph{89.4} & 72.9 & 88.2 & 77.1 & 83.3 \\
        $\mathcal L_{\text{expl}}$ (I) & 69.2 & 57.4 & \cellcolor[HTML]{C0C0C0}\emph{84.3} & 81.0 & 65.2 & 66.1 \\
        \bottomrule
    \end{tabular}
    \caption{Average \textbf{in-domain} AUC plausibility scores ($\uparrow$). $\mathcal L_{\text{expl}}$ (A/R/I) are trained with $\mathcal{L} = \mathcal{L}_{\mathrm{expl}}$, for \textsc{Att}, \textsc{AttR}, and \textsc{IxG} respectively. \colorbox[HTML]{C0C0C0}{\emph{Highlighted values}} correspond to the ER guided attribution technique.}
    \label{tab:sa_plausibility_auc_id}
\end{table}
\begin{table}[t]
    \centering
    \small
    \begin{tabular}{l ccc @{\hskip 1.5em} ccc}
        \toprule
        & \multicolumn{3}{c}{\textbf{Movies}} & \multicolumn{3}{c}{\textbf{Yelp-50}} \\ \cmidrule(r{1em}){2-4} \cmidrule{5-7}
        & \rot{90}{\textsc{Att}} & \rot{90}{\textsc{AttR}} & \rot{90}{\textsc{IxG}} & \rot{90}{\textsc{Att}} & \rot{90}{\textsc{AttR}} & \rot{90}{\textsc{IxG}} \\
        \midrule
        \textsc{Baseline} & 70.0  & 49.6  & 58.1 & 51.9 & 37.1 & 51.5 \\
        \midrule
        \multicolumn{7}{c}{\emph{Joint ER} (\S\ref{subsec:joint_optimisation})} \\
        \midrule
        \textsc{w/ Att} & \cellcolor[HTML]{C0C0C0}\emph{74.6} & 49.1 & 58.1 & \cellcolor[HTML]{C0C0C0}\emph{69.7} & 36.4 & 50.7 \\
        \textsc{w/ AttR} & 70.1 & \cellcolor[HTML]{C0C0C0}\emph{50.2} & 58.0 & 52.5 & \cellcolor[HTML]{C0C0C0}\emph{37.6} & 51.7 \\
        \textsc{w/ IxG} & 69.9 & 49.0 & \cellcolor[HTML]{C0C0C0}\emph{59.5} & 56.5 & 37.3 & \cellcolor[HTML]{C0C0C0}\emph{53.6} \\
        \midrule
        \multicolumn{7}{c}{\emph{Constrained ER} (\S\ref{subsec:constrained_optimisation})} \\
        \midrule
        \textsc{w/ Att} & \cellcolor[HTML]{C0C0C0}\emph{68.8} & 48.1 & 56.0 & \cellcolor[HTML]{C0C0C0}\emph{76.0} & 37.4 & 51.3 \\
        \textsc{w/ AttR} & 71.1 & \cellcolor[HTML]{C0C0C0}\emph{62.8} & 59.3 & 67.2 & \cellcolor[HTML]{C0C0C0}\emph{69.4} & 58.8 \\
        \textsc{w/ IxG} & 56.7 & 56.1 & \cellcolor[HTML]{C0C0C0}\emph{61.6} & 60.8 & 50.5 & \cellcolor[HTML]{C0C0C0}\emph{69.8} \\
        \midrule
        \multicolumn{7}{c}{\emph{$\mathcal{L} = \mathcal{L}_{\mathrm{expl}}$} (\S\ref{subsec:joint_optimisation})} \\
        \midrule
        $\mathcal L_{\text{expl}}$ (A) & \cellcolor[HTML]{C0C0C0}\emph{61.0} & 48.2 & 48.0 & \cellcolor[HTML]{C0C0C0}\emph{70.1} & 38.5 & 47.1 \\
        $\mathcal L_{\text{expl}}$ (R) & 59.4 & \cellcolor[HTML]{C0C0C0}\emph{62.5} & 55.2 & 62.6 & \cellcolor[HTML]{C0C0C0}\emph{68.6} & 56.8 \\
        $\mathcal L_{\text{expl}}$ (I) & 51.8 & 56.9 & \cellcolor[HTML]{C0C0C0}\emph{59.8} & 55.2 & 58.3 & \cellcolor[HTML]{C0C0C0}\emph{68.6} \\
        \bottomrule
    \end{tabular}
    \caption{Average \textbf{out-of-domain} AUC plausibility scores ($\uparrow$). $\mathcal L_{\text{expl}}$ (A/R/I) are trained with $\mathcal{L} = \mathcal{L}_{\mathrm{expl}}$, for \textsc{Att}, \textsc{AttR}, and \textsc{IxG} respectively. \colorbox[HTML]{C0C0C0}{\emph{Highlighted values}} correspond to the ER guided attribution technique.}
    \label{tab:sa_plausibility_auc_ood}
\end{table}

The lack of impact of global guidance on the plausibility of the corresponding guided attributions might come as a surprise. However, we need to consider that: \emph{(i)} this class of  techniques incorporates more of the model's components when computing attributions, potentially making it more difficult for the model to `hide' implausible computations from the explanation loss; and \emph{(ii)} joint ER balances both classification and explanation losses with a preference (via model selection) for classification, meaning that to maintain classification performance the model might require `under-optimising' the explanation loss.
Both possibilities will be further studied in this section.

\begin{figure}[t]
    \centering
    \includegraphics[width=1.0\linewidth]{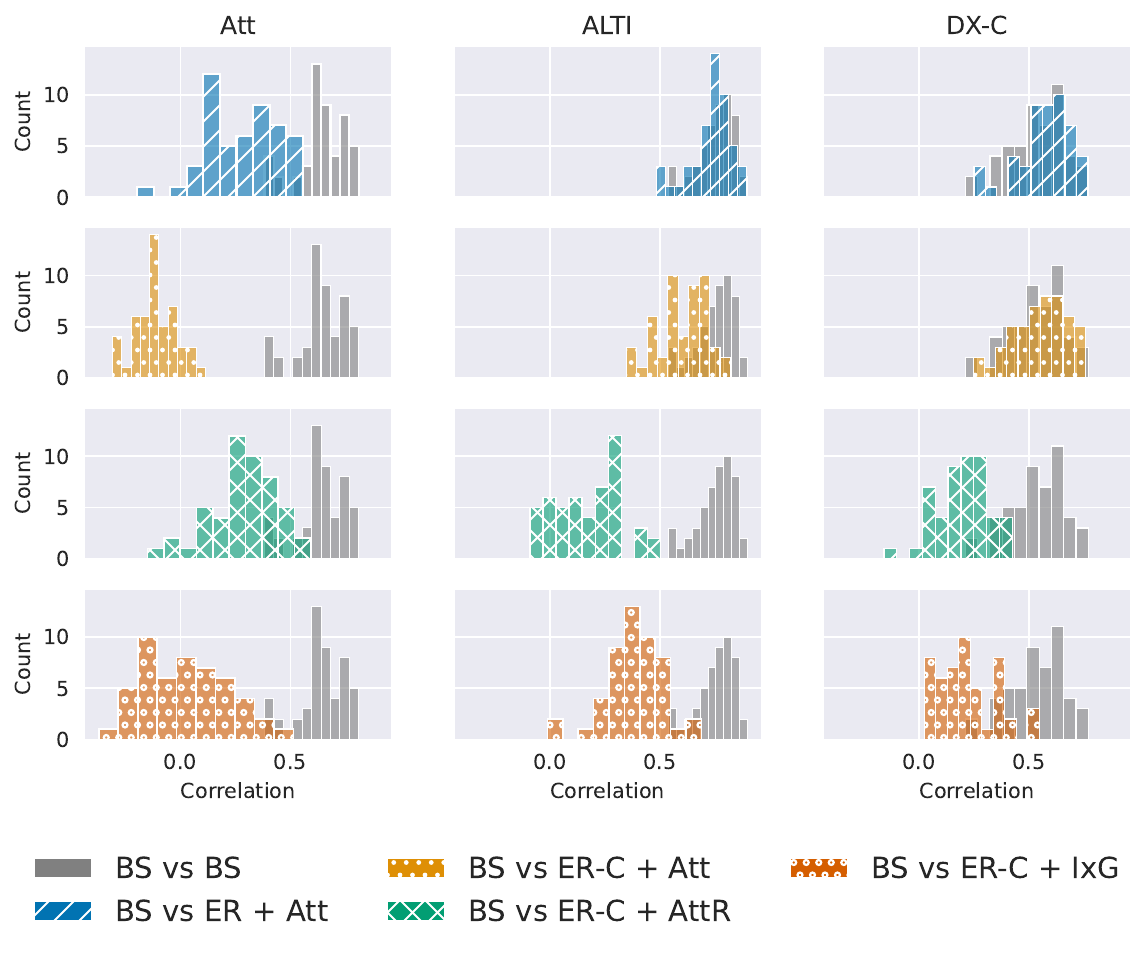}
    \caption{\textbf{Yelp-50} (OOD) Kendall Rank correlations between attribution techniques for different approaches. Baseline (BS) vs Baseline serves as ground-truth for the expected correlations agreement due to seed variability.}
    \label{subfig:sa_attributions_correlations_approaches_yelp}
\end{figure}

\begin{figure}[t]
    \centering
        \centering
        \includegraphics[width=0.95\linewidth]{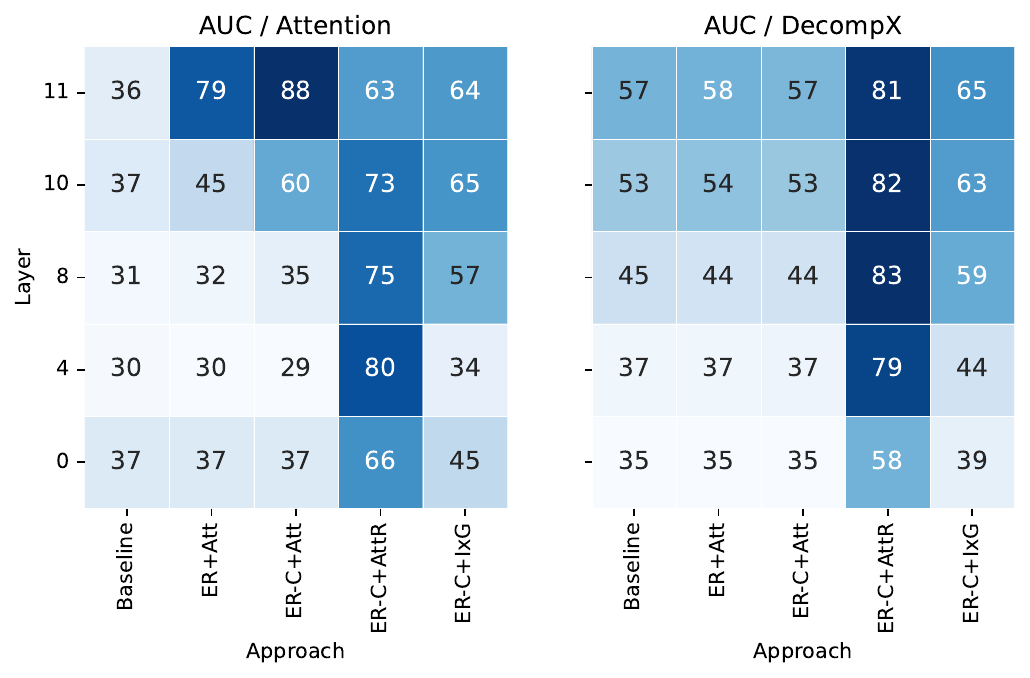}
    \caption{\textbf{SST-Dev} average AUC plausibility score per layer ($\uparrow$) with Attention and DecompX.}
\label{fig:sa_auc_per_layer_attention_decompx_smaller}
\end{figure}

\paragraph{\textsc{ER+Att} fails to impact non-guided attribution techniques.} 
The previous results illustrate how attention guidance (\textsc{ER+Att}) is able to simultaneously improve OOD classification and rationale extraction (when assessed with the corresponding guided attribution technique).
It is tempting to connect these two observations and conclude that \textsc{Att} guidance results in ER models that better classify \emph{while} relying more on plausible tokens. However, we find that not to be the case.
Firstly, we correlate \emph{input attributions} (obtained by a given technique) across training conditions (\eg, ALTI baseline attributions vs. ALTI \textsc{ER+Att} attributions). We find that -- aside from  the guided attribution technique -- correlation coefficients are mostly unaffected by ER training, both in- and out-of-domain (see first row of Figure \ref{subfig:sa_attributions_correlations_approaches_yelp}, for OOD).
We find similar evidence for all non-guided techniques (Appendix Fig. \ref{fig:sa_attributions_correlations_approaches_all}).
Secondly, we compare the AUC plausibility scores of the non-guided  techniques---any non-highlighted cell in Tables \ref{tab:sa_plausibility_auc_id} and \ref{tab:sa_plausibility_auc_ood}---with the corresponding baselines.
Here, we observe a lack of impact across joint ER models. For example, for \textsc{ER+Att} we find the impact to be small (59.0 vs 64.5 for ALTI) to none (58.5 to 58.4 for DX-C). We find similar evidence using average precision and recall@k (Appendix Table \ref{tab:sa_plausibility_all}).
In fact, if we look at AUC plausibility scores across layers (Fig. \ref{fig:sa_auc_per_layer_attention_decompx_smaller}) we can observe that \textsc{ER+Att} only impacts the plausibility of attributions at the top-layer.
This result confirms our suspicions: local  guidance is able to `hide' implausible computations at lower layers, thus hiding them from $\mathcal{L}_{\mathrm{expl}}$.

\begin{figure}[t]
    \centering
        \centering
        \includegraphics[width=0.92\linewidth]{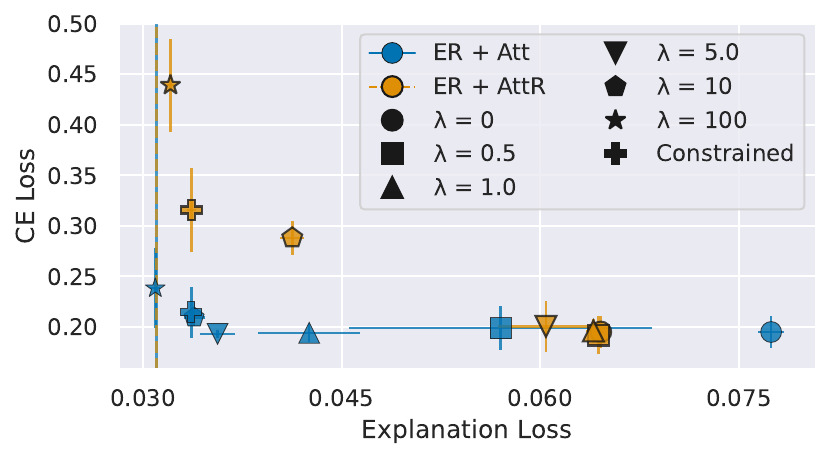}
    \caption{\textbf{SST-Dev} $\mathcal L_{\text{ce}}$ vs. $\mathcal L_{\text{expl}}$. The vertical lines show the validation explanation-loss bounds. Each point is the average of 5 runs, the error bars show standard deviation for each loss. $\lambda$ ranges from 0 (No-ER) to 100.}
\label{fig:sa_multiple-lambdas}
\end{figure}

\paragraph{$\mathcal{L}_{\mathrm{cls}}$ and $\mathcal{L}_{\mathrm{expl}}$ exhibit no synergy.}
We observed before how \textsc{ER+AttR} and \textsc{ER+IxG} failed to impact input attributions, including for the respective guided attribution technique.
To understand why this happened, we train ER models exclusively with the explanation loss, $\mathcal{L} = \mathcal{L}_{\mathrm{expl}}$, and obtain a `lower-bound' for its value.
As seen in Table \ref{tab:sa_losses}, all joint ER approaches exhibit a gap to the explanation loss `lower-bound' value, particularly \textsc{AttR} and \textsc{IxG}, with this observation extending to AUC --- \eg, for $\mathcal{L}_{\text{expl}}$ (A) we obtain a AUC value of 89.0 (vs the 78.6 obtained with joint \textsc{ER+Att}), and for $\mathcal{L}_{\text{expl}}$ (R) we obtain a value of 89.4 (vs the 28.7 obtained with joint \textsc{ER+AttR}).
\begin{table}[t]
    \centering
    \small
    \begin{tabular}{llcc}
        \toprule
        $E_c$ & Objective  & $\mathcal{L}^{(\text{dev})}_{\mathrm{expl}}$ ~($\downarrow$)  &  AUC$^{(\text{dev})}$~($\uparrow$) \\ \midrule        
        \multirow{3}{*}{\textsc{Att}} & $\mathcal L_{\text{expl}}$ & 0.030 & 89.0 \\ 
           & (J) ER  & 0.040 & 78.6 \\
           & (C) ER  & 0.034 & 87.6 \\
        \midrule
        \multirow{3}{*}{\textsc{AttR}} & $\mathcal L_{\text{expl}}$ & 0.032 & 89.4 \\ 
           & (J) ER  & 0.064 & 28.7 \\ 
           & (C) ER  & 0.033 & 89.1 \\
        \midrule
        \multirow{3}{*}{\textsc{IxG}} & $\mathcal L_{\text{expl}}$ & 0.349 & 84.3 \\ 
           & (J) ER  & 0.432 & 53.8 \\ 
           & (C) ER  & 0.358 & 81.5 \\
        \bottomrule         
    \end{tabular}
    \caption{\textbf{SST-Dev} explanation loss and AUC values for models guided with \textsc{Att}, \textsc{AttR}, or \textsc{IxG}, as a function of their training objective (explanation loss alone, vs. (J)oint ER and (C)onstrained ER).}
    \label{tab:sa_losses}
\end{table}
These results highlight one possible difficulty of the joint ER setup: promoting plausible attributions via ER with global guidance might incur a negative cost in classification performance, hence joint optimisation (whose hyperparameters are selected based on classification loss) will under-optimise the auxiliary task.
In order to confirm this behavior we run the joint ER setup with increasing values of $\lambda$. The results for \textsc{ER+Att} and \textsc{ER+AttR} can be seen in Fig. \ref{fig:sa_multiple-lambdas}.
We observe two overall trends.
Given a high enough $\lambda$, the explanation loss converges to the obtained `lower-bound'. For \textsc{ER+Att}, even smaller values of $\lambda$ impact the explanation loss, whereas for \textsc{ER+AttR} larger values are required.
For both strategies alike, as the explanation loss values decreases, the cross entropy loss increases (we verify that all models still converge on the classification loss).
However, the \textsc{ER+Att} model seems to be more robust to this trade-off---we observe a smaller increase in CE loss as the explanation loss decreases when compared to \textsc{ER+AttR}.
We observe a similar effect for \textsc{ER+IxG} in Appendix Fig. \ref{fig:sa_multiple-lambdas_ixg}.\looseness=-1

Although this is, perhaps, not a necessarily surprising finding (\citet{carton2022learn}, for example, observe a similar tradeoff, though in a multi-tasking setting), it challenges a key assumption of ER as a means to improve OOD generalisation: that the explanation loss should help inform the classifier, improving OOD performance.
In fact, we observe that the joint ER setup will struggle to accommodate both losses, particularly when using a guided attribution technique that is more difficult to `hack' in order to become apparently better at predicting human rationales, such as attention-rollout, where the explanation loss might be lowered, but far from optimally.\looseness=-1

\subsection{Constrained Optimisation}
\label{subsec:constrained_optimisation}

In \S \ref{subsec:joint_optimisation}, we observed how joint ER under-optimises the explanation loss and how this effect varies in function of the guided attribution technique being local or global.
Moreover, Table \ref{tab:sa_plausibility_auc_id} shows that optimising exclusively the explanation loss, $\mathcal{L} = \mathcal{L}_{\mathrm{expl}}$, can lead to AUC plausibility improvements---on the guided attribution technique alone for local guidance, and across the board for global guidance.
This serves as motivation to analyse classifiers optimised for the classification task, while subject to achieving the aforementioned explanation loss `lower-bound' via constrained optimisation.

\paragraph{Constrained global guidance is necessary to impact non-guided attribution techniques.}
We start by noting how constrained ER addresses an earlier result: both global guided techniques are now impacting the corresponding guided attribution technique.
This is true for both the AUC plausibility scores (in Table \ref{tab:sa_plausibility_auc_id}) and correlation coefficients (in Appendix Figure \ref{fig:sa_attributions_correlations_approaches_all}).
However, this is so by design, and a more interesting observation stems from inspecting the non-guided attribution techniques --- while constrained \textsc{ER-C+Att} still mostly fails to impact other attribution techniques, constrained \textsc{ER-C+AttR} and \textsc{ER-C+IxG} impact the outcome of all assessed attribution techniques.
We can observe this along three `dimensions'.
First, global techniques are able to clearly impact the plausibility of all attribution techniques (Table \ref{tab:sa_plausibility_auc_id}). For instance, \textsc{ER-C+AttR} leads to large improvements in DecompX, where the average AUC plausibility score is improved from 57.5 (in baseline) to 81.3, with the value being 56.8 for \textsc{ER-C+Att}.
Second, there is a clear impact in the correlation of attributions across approaches (Appendix Fig. \ref{subfig:sa_attributions_correlations_approaches_all_sst-dev}), contrarily to what is observed for local ER, including \textsc{ER-C+Att}.
Finally, AUC plausibility scores per layer (Figure \ref{fig:sa_auc_per_layer_attention_decompx_smaller}) also show how the global technique is able to impact not only attributions at the top-layer, but across the whole model, making them more aligned with the human rationales.
All observations along the three inspected `dimensions' hold for the OOD data, with attributions clearly being impacted (Figure \ref{subfig:sa_attributions_correlations_approaches_yelp}), despite the lower influence in the AUC plausibility metric (Table \ref{tab:sa_plausibility_auc_ood}).\footnote{We also find similar evidence for AUC per-layer across multiple attribution techniques (Appendix Figure \ref{fig:appendix_auc_per_layer_all}).}

\paragraph{There is a disconnect between (over-)reliance on plausible features and OOD robustness.}
Despite the noticeable impact of constrained \textsc{ER-C+AttR} and \textsc{ER-C+IxG} on attributions and their plausibility, Table \ref{tab:sa_results} indicates a decline in classification performance.
That is, the only ER strategies that resulted in models that rely more strongly on plausible tokens decrease OOD classification robustness.
This is somewhat expected given what we observed before in Figure \ref{fig:sa_multiple-lambdas}. However, it does lead us to conclude that the ER protocol should be re-considered, or at least, a bigger emphasis on trying to understand what `amount' of increased plausibility is desirable, and where that information should be used while regularising the model, in order to better align the current expectations of improved OOD generalisation due to an increased reliance of ER models on plausible features.

\subsection{Predicting OOD performance}
\label{subsec:predict_ood_performance}
    
As shown in Figure \ref{fig:sa_results}, ER models exhibit a broad spread of OOD classification results. This highlights the importance of identifying the best-performing models, ideally without access to OOD data. Thus, we investigate whether it is possible to predict OOD classification performance from features potentially available for model selection.

\begin{figure}[t]
    \centering
    \begin{subfigure}[t]{0.49\linewidth}
        \centering
        \includegraphics[width=\linewidth]{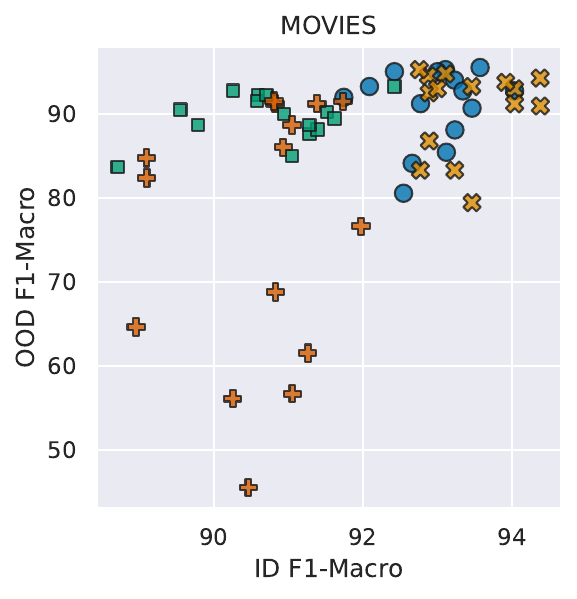}
        \caption{ID Classification}
        \label{subfig:sa_ood_prediction_cls}
    \end{subfigure}
    \hfill
    \begin{subfigure}[t]{0.49\linewidth}
        \centering
        \includegraphics[width=\linewidth]{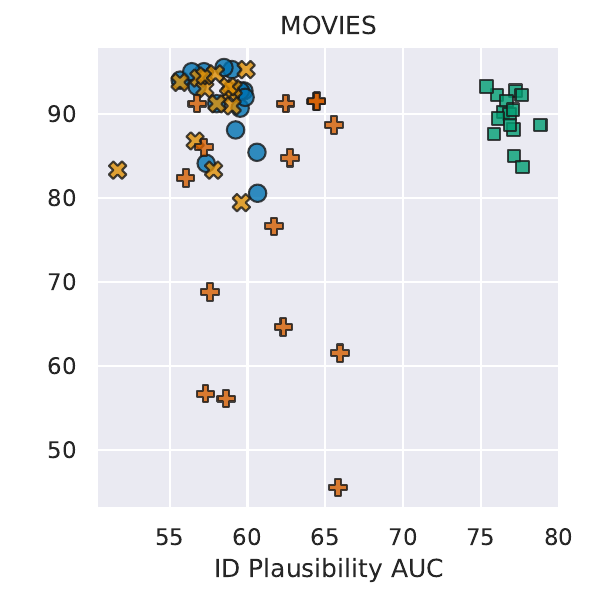}
        \caption{ID Plaus. AUC w/ \textsc{DX-C}}
        \label{subfig:sa_ood_prediction_auc}
    \end{subfigure}
    \vskip\baselineskip
    \begin{subfigure}[t]{0.98\linewidth}
        \centering
        \includegraphics[width=\linewidth]{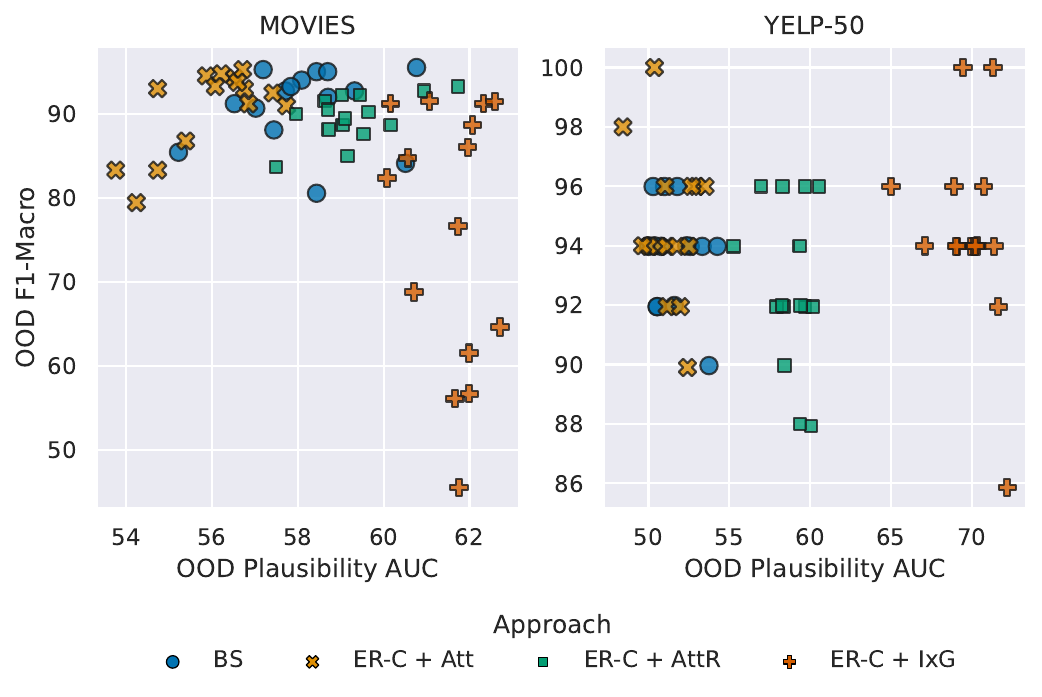}
        \caption{OOD Plausibility AUC w/ \textsc{IxG}}
        \label{subfig:sa_ood_prediction_ood_auc}
    \end{subfigure}
    \caption{Relationship between ID/OOD predictors and OOD classification performance.}
\label{fig:sa_ood_prediction}
\end{figure}

\paragraph{In-domain classification and plausibility performance are not predictive of OOD improvements.}

Figures \ref{subfig:sa_ood_prediction_cls} and \ref{subfig:sa_ood_prediction_auc} show the association between in-domain (SST-Dev) predictors classification F1-Macro and plausibility AUC (computed with \textsc{DX-C}), and classification F1-Macro for Movies (OOD).
In both cases there is no clear correlation between the in-domain metrics and OOD classification scores, with the result in Figure \ref{subfig:sa_ood_prediction_auc} further supporting the apparent disconnect between reliance on plausible features and OOD classification performance.
We find a similar behavior for other approaches, OOD datasets, plausibility metrics and attribution techniques (Appendix \ref{sec_appendix:predicting_ood_performance}).

\paragraph{OOD plausibility performance is not predictive of OOD improvements.}

Given that we do not find in-domain classification performance or attribution plausibility to be predictive of OOD performance, we now investigate if there is any correlation at the OOD level.
In order to test this, we use the Movies and Yelp-50 datasets.
Figure \ref{subfig:sa_ood_prediction_ood_auc} shows the association between OOD plausibility AUC (computed with IxG\footnote{\textsc{Movies} consists of long inputs, resulting in prohibitive runtime for batch sizes that do not result in OOM errors, on a Nvidia A100, when using DX and ALTI. Thus, we report IxG, as it can be computed for both datasets in the figure.}) and classification F1-Macro, for both datasets. Similarly to what we observed in-domain, there is no correlation between OOD plausibility scores and its corresponding classification task performance.

The combination of both these results highlights  the difficulty of predicting OOD results, and brings forth another challenge of the ER setup, where we need to make decisions about design choices based on in-domain performance, that does not seem to be linked to OOD generalisation.

\section{Conclusion}

In this work we take a step towards a better understanding of how explanation regularisation impacts a model beyond its classification performance. In particular, we aim to study both in- and out-of-domain settings, with a focus on how the plausibility of input attributions is affected, and what is the relationship between increased reliance on plausible tokens and robustness to OOD conditions.

We find that ER, unless constrained to meet a `lower-bound' on the explanation loss computed with a global attribution technique, does not lead to models that effectively rely on plausible tokens.
In fact, we observe a disconnect between reliance on plausible tokens and OOD robustness, with \emph{(i)} OOD classification improvements not being predicted by increased  plausibility; and \emph{(ii)} OOD classification performance degrading when models are constrained to rely more on plausible tokens.

Our findings highlight relevant challenges for ER and suggest future research questions: \emph{(i)} how to achieve a better balance between solving the classification task and constraining the model to attend to plausible tokens, \eg, by learning which layers and attention heads of the Transformer model to regularise (see \citet{fernandes2022scaffold}), or \emph{(ii)} by allowing the model to learn to select which rationales to use during training (see \citet{arous2021marta,carton2022learn}); and \emph{(iii)} how would input attributions for an ER model be impacted by using more informed, while still efficient, vector-based global guidance techniques as part of ER design choices (and not only as an analysis tool).

\section*{Limitations}

In this section we identify some limitations of our work. However, we do note that we focus on the limitations that stem from design choices of our analysis, and not on the limitations we `inherit' and that are part of the base problem, \eg, whether plausibility should be a desired property of explanations \citep{jacovi2021aligning}.

\paragraph{Data.} In terms of data, we identify two limitations.
The first limitation related with data concerns the limited number of datasets per task that include annotations, making it difficult to analyse the impact of ER on the attributions on out-of-domain settings.
Secondly, we use only the task of sentiment analysis to study the disconnect between ER's robustness to OOD conditions and the plausibility of attributions. This is also connected to the first discussed limitation --- a full ER setup requires rationale-annotated data of an appreciable dimension for training, and also multiple OOD datasets, including at least one with rationale-annotated data.
However, the questions we ask are deeply inherent to ER, and it seems unlikely that alternative tasks would be impacted differently.

\paragraph{Models.} We use a single pre-trained model, \textsc{BigBird-roberta-base} \citep{zaheer2020big}. This choice follows existing work on ER (in particular applied to out-of-domain conditions), and allows evaluating datasets with long examples, but it does mean that we do not assess how ER impacts attributions for different pre-trained models, in particular, with varying number of parameters.

\paragraph{Attribution Techniques.} Two of the techniques we use, ALTI \citep{ferrando2022measuring} and DecompX \citep{modarressi2023decompx}, cannot be applied to long examples, leading either to prohibitive runtime, or to out-of-memory errors (on a single GPU NVIDIA A100), even with a batch size as low as 2. Thus, we cannot apply them to the OOD movies dataset, and have to limit our manually annotated split of Yelp to a moderate maximum sequence length. Yet, on Yelp, these represent a reasonable portion of the dataset's length distribution.

\section*{Ethical Considerations}

This work studies the synergy between models that attribute more plausibly, \ie, more aligned with humans, and improved OOD performance. We use existing datasets, that include human annotations of the relevant tokens that explain why an example is classified with a given label. These annotations might be biased, include mistakes, etc. By training a text classifier to become more aligned with those annotations we might further magnify the biases of the annotation process.

\section*{Acknowledgements}

This research was done within the Mercury Machine Learning Lab, a collaboration between the University of Amsterdam, TU Delft, and Booking.com.
Ivan Titov is supported by the Dutch National Science Foundation (NWO Vici VI.C.212.053).
All content represents the opinion of the authors, which is not necessarily shared or endorsed by their respective employers and/or sponsors.

\bibliography{custom}

\begin{thebibliography}{61}
\providecommand{\natexlab}[1]{#1}

\bibitem[{Abnar and Zuidema(2020)}]{abnar2020quantifying}
Samira Abnar and Willem Zuidema. 2020.
\newblock Quantifying attention flow in transformers.
\newblock In \emph{Proceedings of the 58th Annual Meeting of the Association for Computational Linguistics}, pages 4190--4197.

\bibitem[{AI@Meta(2024)}]{llama3modelcard}
AI@Meta. 2024.
\newblock \href {https://github.com/meta-llama/llama3/blob/main/MODEL_CARD.md} {Llama 3 model card}.

\bibitem[{Arous et~al.(2021)Arous, Dolamic, Yang, Bhardwaj, Cuccu, and Cudr{\'e}-Mauroux}]{arous2021marta}
Ines Arous, Ljiljana Dolamic, Jie Yang, Akansha Bhardwaj, Giuseppe Cuccu, and Philippe Cudr{\'e}-Mauroux. 2021.
\newblock Marta: Leveraging human rationales for explainable text classification.
\newblock In \emph{Proceedings of the AAAI conference on artificial intelligence}, volume~35, pages 5868--5876.

\bibitem[{Boyd and Vandenberghe(2004)}]{boyd2004convex}
Stephen~P Boyd and Lieven Vandenberghe. 2004.
\newblock \emph{Convex optimization}.
\newblock Cambridge university press.

\bibitem[{Camburu et~al.(2018)Camburu, Rockt{\"a}schel, Lukasiewicz, and Blunsom}]{camburu2018snli}
Oana-Maria Camburu, Tim Rockt{\"a}schel, Thomas Lukasiewicz, and Phil Blunsom. 2018.
\newblock e-snli: Natural language inference with natural language explanations.
\newblock \emph{Advances in Neural Information Processing Systems}, 31.

\bibitem[{Carton et~al.(2022)Carton, Kanoria, and Tan}]{carton2022learn}
Samuel Carton, Surya Kanoria, and Chenhao Tan. 2022.
\newblock What to learn, and how: Toward effective learning from rationales.
\newblock In \emph{Findings of the Association for Computational Linguistics: ACL 2022}, pages 1075--1088.

\bibitem[{Carton et~al.(2020)Carton, Rathore, and Tan}]{carton2020evaluating}
Samuel Carton, Anirudh Rathore, and Chenhao Tan. 2020.
\newblock Evaluating and characterizing human rationales.
\newblock In \emph{Proceedings of the 2020 Conference on Empirical Methods in Natural Language Processing (EMNLP)}, pages 9294--9307.

\bibitem[{Chan et~al.(2022)Chan, Sanjabi, Mathias, Tan, Nie, Peng, Ren, and Firooz}]{chan2022unirex}
Aaron Chan, Maziar Sanjabi, Lambert Mathias, Liang Tan, Shaoliang Nie, Xiaochang Peng, Xiang Ren, and Hamed Firooz. 2022.
\newblock Unirex: A unified learning framework for language model rationale extraction.
\newblock In \emph{International Conference on Machine Learning}, pages 2867--2889. PMLR.

\bibitem[{Chefer et~al.(2021)Chefer, Gur, and Wolf}]{chefer2021transformer}
Hila Chefer, Shir Gur, and Lior Wolf. 2021.
\newblock Transformer interpretability beyond attention visualization.
\newblock In \emph{Proceedings of the IEEE/CVF conference on computer vision and pattern recognition}, pages 782--791.

\bibitem[{Chefer et~al.(2022)Chefer, Schwartz, and Wolf}]{chefer2022robustness}
Hila Chefer, Idan Schwartz, and Lior Wolf. 2022.
\newblock Optimizing relevance maps of vision transformers improves robustness.
\newblock \emph{Advances in Neural Information Processing Systems}.

\bibitem[{Clark et~al.(2019)Clark, Khandelwal, Levy, and Manning}]{clark-etal-2019-bert}
Kevin Clark, Urvashi Khandelwal, Omer Levy, and Christopher~D. Manning. 2019.
\newblock \href {https://doi.org/10.18653/v1/W19-4828} {What does {BERT} look at? an analysis of {BERT}{'}s attention}.
\newblock In \emph{Proceedings of the 2019 ACL Workshop BlackboxNLP: Analyzing and Interpreting Neural Networks for NLP}, pages 276--286, Florence, Italy. Association for Computational Linguistics.

\bibitem[{DeYoung et~al.(2020)DeYoung, Jain, Rajani, Lehman, Xiong, Socher, and Wallace}]{deyoung2020eraser}
Jay DeYoung, Sarthak Jain, Nazneen~Fatema Rajani, Eric Lehman, Caiming Xiong, Richard Socher, and Byron~C Wallace. 2020.
\newblock Eraser: A benchmark to evaluate rationalized nlp models.
\newblock In \emph{Proceedings of the 58th Annual Meeting of the Association for Computational Linguistics}, pages 4443--4458.

\bibitem[{Falcon and {The PyTorch Lightning team}(2019)}]{Falcon_PyTorch_Lightning_2019}
William Falcon and {The PyTorch Lightning team}. 2019.
\newblock \href {https://doi.org/10.5281/zenodo.3828935} {{PyTorch Lightning}}.

\bibitem[{Fernandes et~al.(2022)Fernandes, Treviso, Pruthi, Martins, and Neubig}]{fernandes2022scaffold}
Patrick Fernandes, Marcos Treviso, Danish Pruthi, André~F.T. Martins, and Graham Neubig. 2022.
\newblock Learning to scaffold: Optimizing model explanations for teaching.
\newblock In \emph{Advances in Neural Information Processing Systems}.

\bibitem[{Ferrando et~al.(2022)Ferrando, G{\'a}llego, and Costa-juss{\`a}}]{ferrando2022measuring}
Javier Ferrando, Gerard~I G{\'a}llego, and Marta~R Costa-juss{\`a}. 2022.
\newblock Measuring the mixing of contextual information in the transformer.
\newblock In \emph{Proceedings of the 2022 Conference on Empirical Methods in Natural Language Processing}, pages 8698--8714.

\bibitem[{Fomicheva et~al.(2021)Fomicheva, Lertvittayakumjorn, Zhao, Eger, and Gao}]{fomicheva2021eval4nlp}
Marina Fomicheva, Piyawat Lertvittayakumjorn, Wei Zhao, Steffen Eger, and Yang Gao. 2021.
\newblock The eval4nlp shared task on explainable quality estimation: Overview and results.
\newblock In \emph{Proceedings of the 2nd Workshop on Evaluation and Comparison of NLP Systems}, pages 165--178.

\bibitem[{Fomicheva et~al.(2022)Fomicheva, Specia, and Aletras}]{fomicheva2022translation}
Marina Fomicheva, Lucia Specia, and Nikolaos Aletras. 2022.
\newblock Translation error detection as rationale extraction.
\newblock In \emph{Findings of the Association for Computational Linguistics: ACL 2022}, pages 4148--4159.

\bibitem[{Ghaeini et~al.(2019)Ghaeini, Fern, Shahbazi, and Tadepalli}]{ghaeini2019saliency}
Reza Ghaeini, Xiaoli~Z Fern, Hamed Shahbazi, and Prasad Tadepalli. 2019.
\newblock Saliency learning: Teaching the model where to pay attention.
\newblock In \emph{Proceedings of NAACL-HLT}, pages 4016--4025.

\bibitem[{Hartmann and Sonntag(2022)}]{hartmann2022survey}
Mareike Hartmann and Daniel Sonntag. 2022.
\newblock A survey on improving nlp models with human explanations.
\newblock In \emph{Proceedings of the First Workshop on Learning with Natural Language Supervision}, pages 40--47.

\bibitem[{Hase and Bansal(2022)}]{hase-bansal-2022-models}
Peter Hase and Mohit Bansal. 2022.
\newblock \href {https://doi.org/10.18653/v1/2022.lnls-1.4} {When can models learn from explanations? a formal framework for understanding the roles of explanation data}.
\newblock In \emph{Proceedings of the First Workshop on Learning with Natural Language Supervision}, pages 29--39, Dublin, Ireland. Association for Computational Linguistics.

\bibitem[{Honnibal et~al.(2020)Honnibal, Montani, Van~Landeghem, and Boyd}]{spacy2020}
Matthew Honnibal, Ines Montani, Sofie Van~Landeghem, and Adriane Boyd. 2020.
\newblock \href {https://doi.org/10.5281/zenodo.1212303} {spacy: Industrial-strength natural language processing in python}.

\bibitem[{Hou et~al.(2024)Hou, Li, He, Yan, Chen, and McAuley}]{hou2024bridging}
Yupeng Hou, Jiacheng Li, Zhankui He, An~Yan, Xiusi Chen, and Julian McAuley. 2024.
\newblock Bridging language and items for retrieval and recommendation.
\newblock \emph{arXiv preprint arXiv:2403.03952}.

\bibitem[{Ismail et~al.(2021)Ismail, Corrada~Bravo, and Feizi}]{ismail2021improving}
Aya~Abdelsalam Ismail, Hector Corrada~Bravo, and Soheil Feizi. 2021.
\newblock Improving deep learning interpretability by saliency guided training.
\newblock \emph{Advances in Neural Information Processing Systems}, 34:26726--26739.

\bibitem[{Jacovi and Goldberg(2021)}]{jacovi2021aligning}
Alon Jacovi and Yoav Goldberg. 2021.
\newblock \href {https://doi.org/10.1162/tacl\_a\_00367} {{Aligning Faithful Interpretations with their Social Attribution}}.
\newblock \emph{Transactions of the Association for Computational Linguistics}, 9:294--310.

\bibitem[{Jain and Wallace(2019)}]{jain2019attention}
Sarthak Jain and Byron~C Wallace. 2019.
\newblock Attention is not explanation.
\newblock In \emph{Proceedings of the 2019 Conference of the North American Chapter of the Association for Computational Linguistics: Human Language Technologies, Volume 1 (Long and Short Papers)}, pages 3543--3556.

\bibitem[{Joshi et~al.(2022)Joshi, Chan, Liu, Nie, Sanjabi, Firooz, and Ren}]{joshi-etal-2022-er}
Brihi Joshi, Aaron Chan, Ziyi Liu, Shaoliang Nie, Maziar Sanjabi, Hamed Firooz, and Xiang Ren. 2022.
\newblock \href {https://aclanthology.org/2022.findings-emnlp.242} {{ER}-test: Evaluating explanation regularization methods for language models}.
\newblock In \emph{Findings of the Association for Computational Linguistics: EMNLP 2022}, pages 3315--3336, Abu Dhabi, United Arab Emirates. Association for Computational Linguistics.

\bibitem[{Kindermans et~al.(2016)Kindermans, Sch{\"u}tt, M{\"u}ller, and D{\"a}hne}]{kindermans2016investigating}
Pieter-Jan Kindermans, Kristof Sch{\"u}tt, Klaus-Robert M{\"u}ller, and Sven D{\"a}hne. 2016.
\newblock Investigating the influence of noise and distractors on the interpretation of neural networks.
\newblock \emph{arXiv preprint arXiv:1611.07270}.

\bibitem[{Kobayashi et~al.(2020)Kobayashi, Kuribayashi, Yokoi, and Inui}]{kobayashi-etal-2020-attention}
Goro Kobayashi, Tatsuki Kuribayashi, Sho Yokoi, and Kentaro Inui. 2020.
\newblock \href {https://doi.org/10.18653/v1/2020.emnlp-main.574} {Attention is not only a weight: Analyzing transformers with vector norms}.
\newblock In \emph{Proceedings of the 2020 Conference on Empirical Methods in Natural Language Processing (EMNLP)}, pages 7057--7075, Online. Association for Computational Linguistics.

\bibitem[{Kobayashi et~al.(2021)Kobayashi, Kuribayashi, Yokoi, and Inui}]{kobayashi-etal-2021-incorporating}
Goro Kobayashi, Tatsuki Kuribayashi, Sho Yokoi, and Kentaro Inui. 2021.
\newblock \href {https://doi.org/10.18653/v1/2021.emnlp-main.373} {{I}ncorporating {R}esidual and {N}ormalization {L}ayers into {A}nalysis of {M}asked {L}anguage {M}odels}.
\newblock In \emph{Proceedings of the 2021 Conference on Empirical Methods in Natural Language Processing}, pages 4547--4568, Online and Punta Cana, Dominican Republic. Association for Computational Linguistics.

\bibitem[{Kokhlikyan et~al.(2020)Kokhlikyan, Miglani, Martin, Wang, Alsallakh, Reynolds, Melnikov, Kliushkina, Araya, Yan et~al.}]{kokhlikyan2020captum}
Narine Kokhlikyan, Vivek Miglani, Miguel Martin, Edward Wang, Bilal Alsallakh, Jonathan Reynolds, Alexander Melnikov, Natalia Kliushkina, Carlos Araya, Siqi Yan, et~al. 2020.
\newblock Captum: A unified and generic model interpretability library for pytorch.
\newblock \emph{arXiv preprint arXiv:2009.07896}.

\bibitem[{Liu and Avci(2019)}]{liu2019incorporating}
Frederick Liu and Besim Avci. 2019.
\newblock Incorporating priors with feature attribution on text classification.
\newblock In \emph{Proceedings of the 57th Annual Meeting of the Association for Computational Linguistics}, pages 6274--6283.

\bibitem[{Loshchilov and Hutter(2018)}]{loshchilov2018decoupled}
Ilya Loshchilov and Frank Hutter. 2018.
\newblock Decoupled weight decay regularization.
\newblock In \emph{International Conference on Learning Representations}.

\bibitem[{Maas et~al.(2011)Maas, Daly, Pham, Huang, Ng, and Potts}]{maas2011learningvectors}
Andrew~L. Maas, Raymond~E. Daly, Peter~T. Pham, Dan Huang, Andrew~Y. Ng, and Christopher Potts. 2011.
\newblock \href {http://www.aclweb.org/anthology/P11-1015} {Learning word vectors for sentiment analysis}.
\newblock In \emph{Proceedings of the 49th Annual Meeting of the Association for Computational Linguistics: Human Language Technologies}, pages 142--150, Portland, Oregon, USA. Association for Computational Linguistics.

\bibitem[{Madani and Minervini(2023)}]{madani2023refer}
Mohammad Reza~Ghasemi Madani and Pasquale Minervini. 2023.
\newblock Refer: An end-to-end rationale extraction framework for explanation regularization.
\newblock In \emph{Proceedings of the 27th Conference on Computational Natural Language Learning (CoNLL)}, pages 587--602.

\bibitem[{Mathew et~al.(2021)Mathew, Saha, Yimam, Biemann, Goyal, and Mukherjee}]{mathew2021hatexplain}
Binny Mathew, Punyajoy Saha, Seid~Muhie Yimam, Chris Biemann, Pawan Goyal, and Animesh Mukherjee. 2021.
\newblock Hatexplain: A benchmark dataset for explainable hate speech detection.
\newblock In \emph{Proceedings of the AAAI conference on artificial intelligence}, volume~35, pages 14867--14875.

\bibitem[{Modarressi et~al.(2023)Modarressi, Fayyaz, Aghazadeh, Yaghoobzadeh, and Pilehvar}]{modarressi2023decompx}
Ali Modarressi, Mohsen Fayyaz, Ehsan Aghazadeh, Yadollah Yaghoobzadeh, and Mohammad~Taher Pilehvar. 2023.
\newblock \href {https://doi.org/10.18653/v1/2023.acl-long.149} {{D}ecomp{X}: Explaining transformers decisions by propagating token decomposition}.
\newblock In \emph{Proceedings of the 61st Annual Meeting of the Association for Computational Linguistics (Volume 1: Long Papers)}, pages 2649--2664, Toronto, Canada. Association for Computational Linguistics.

\bibitem[{Modarressi et~al.(2022)Modarressi, Fayyaz, Yaghoobzadeh, and Pilehvar}]{modarressi2022globenc}
Ali Modarressi, Mohsen Fayyaz, Yadollah Yaghoobzadeh, and Mohammad~Taher Pilehvar. 2022.
\newblock Globenc: Quantifying global token attribution by incorporating the whole encoder layer in transformers.
\newblock In \emph{Proceedings of the 2022 Conference of the North American Chapter of the Association for Computational Linguistics: Human Language Technologies}, pages 258--271.

\bibitem[{Paszke et~al.(2019)Paszke, Gross, Massa, Lerer, Bradbury, Chanan, Killeen, Lin, Gimelshein, Antiga et~al.}]{paszke2019pytorch}
Adam Paszke, Sam Gross, Francisco Massa, Adam Lerer, James Bradbury, Gregory Chanan, Trevor Killeen, Zeming Lin, Natalia Gimelshein, Luca Antiga, et~al. 2019.
\newblock Pytorch: An imperative style, high-performance deep learning library.
\newblock \emph{Advances in neural information processing systems}, 32.

\bibitem[{Plumb et~al.(2020)Plumb, Al-Shedivat, Cabrera, Perer, Xing, and Talwalkar}]{plumb2020regularizing}
Gregory Plumb, Maruan Al-Shedivat, {\'A}ngel~Alexander Cabrera, Adam Perer, Eric Xing, and Ameet Talwalkar. 2020.
\newblock Regularizing black-box models for improved interpretability.
\newblock \emph{Advances in Neural Information Processing Systems}, 33:10526--10536.

\bibitem[{Pruthi et~al.(2022)Pruthi, Bansal, Dhingra, Soares, Collins, Lipton, Neubig, and Cohen}]{pruthi2022evaluating}
Danish Pruthi, Rachit Bansal, Bhuwan Dhingra, Livio~Baldini Soares, Michael Collins, Zachary~C Lipton, Graham Neubig, and William~W Cohen. 2022.
\newblock Evaluating explanations: How much do explanations from the teacher aid students?
\newblock \emph{Transactions of the Association for Computational Linguistics}, 10:359--375.

\bibitem[{Pruthi et~al.(2020)Pruthi, Gupta, Dhingra, Neubig, and Lipton}]{pruthi2020learning}
Danish Pruthi, Mansi Gupta, Bhuwan Dhingra, Graham Neubig, and Zachary~C Lipton. 2020.
\newblock Learning to deceive with attention-based explanations.
\newblock In \emph{Proceedings of the 58th Annual Meeting of the Association for Computational Linguistics}, pages 4782--4793.

\bibitem[{Qiang et~al.(2022)Qiang, Pan, Li, Li, Jang, and Zhu}]{qiang2022attcat}
Yao Qiang, Deng Pan, Chengyin Li, Xin Li, Rhongho Jang, and Dongxiao Zhu. 2022.
\newblock Attcat: Explaining transformers via attentive class activation tokens.
\newblock \emph{Advances in neural information processing systems}, 35:5052--5064.

\bibitem[{Rajani et~al.(2019)Rajani, McCann, Xiong, and Socher}]{rajani-etal-2019-explain}
Nazneen~Fatema Rajani, Bryan McCann, Caiming Xiong, and Richard Socher. 2019.
\newblock \href {https://doi.org/10.18653/v1/P19-1487} {Explain yourself! leveraging language models for commonsense reasoning}.
\newblock In \emph{Proceedings of the 57th Annual Meeting of the Association for Computational Linguistics}, pages 4932--4942, Florence, Italy. Association for Computational Linguistics.

\bibitem[{Rao et~al.(2023)Rao, B{\"o}hle, Parchami-Araghi, and Schiele}]{rao2023studying}
Sukrut Rao, Moritz B{\"o}hle, Amin Parchami-Araghi, and Bernt Schiele. 2023.
\newblock Studying how to efficiently and effectively guide models with explanations.
\newblock In \emph{Proceedings of the IEEE/CVF International Conference on Computer Vision}, pages 1922--1933.

\bibitem[{Resck et~al.(2024)Resck, Raimundo, and Poco}]{resck2024exploring}
Lucas~E. Resck, Marcos~M. Raimundo, and Jorge Poco. 2024.
\newblock \href {https://arxiv.org/abs/2404.03098} {Exploring the {Trade}-off {Between} {Model} {Performance} and {Explanation} {Plausibility} of {Text} {Classifiers} {Using} {Human} {Rationales}}.
\newblock In \emph{Findings of the {Association} for {Computational} {Linguistics}: {NAACL} 2024}. Association for Computational Linguistics.

\bibitem[{Rieger et~al.(2020)Rieger, Singh, Murdoch, and Yu}]{rieger2020interpretations}
Laura Rieger, Chandan Singh, William Murdoch, and Bin Yu. 2020.
\newblock Interpretations are useful: penalizing explanations to align neural networks with prior knowledge.
\newblock In \emph{International conference on machine learning}, pages 8116--8126. PMLR.

\bibitem[{Ross et~al.(2017)Ross, Hughes, and Doshi-Velez}]{ross2017rrr}
Andrew~Slavin Ross, Michael~C. Hughes, and Finale Doshi-Velez. 2017.
\newblock \href {https://doi.org/10.24963/ijcai.2017/371} {Right for the right reasons: Training differentiable models by constraining their explanations}.
\newblock In \emph{Proceedings of the Twenty-Sixth International Joint Conference on Artificial Intelligence, {IJCAI-17}}, pages 2662--2670.

\bibitem[{Serrano and Smith(2019)}]{serrano2019attention}
Sofia Serrano and Noah~A Smith. 2019.
\newblock Is attention interpretable?
\newblock In \emph{Proceedings of the 57th Annual Meeting of the Association for Computational Linguistics}, pages 2931--2951.

\bibitem[{Shrikumar et~al.(2017)Shrikumar, Greenside, and Kundaje}]{shrikumar2017learning}
Avanti Shrikumar, Peyton Greenside, and Anshul Kundaje. 2017.
\newblock Learning important features through propagating activation differences.
\newblock In \emph{International conference on machine learning}, pages 3145--3153. PMLR.

\bibitem[{Skalse et~al.(2022)Skalse, Howe, Krasheninnikov, and Krueger}]{skalse2022defining}
Joar Skalse, Nikolaus Howe, Dmitrii Krasheninnikov, and David Krueger. 2022.
\newblock Defining and characterizing reward gaming.
\newblock \emph{Advances in Neural Information Processing Systems}, 35:9460--9471.

\bibitem[{Socher et~al.(2013)Socher, Perelygin, Wu, Chuang, Manning, Ng, and Potts}]{socher-etal-2013-recursive}
Richard Socher, Alex Perelygin, Jean Wu, Jason Chuang, Christopher~D. Manning, Andrew Ng, and Christopher Potts. 2013.
\newblock \href {https://aclanthology.org/D13-1170} {Recursive deep models for semantic compositionality over a sentiment treebank}.
\newblock In \emph{Proceedings of the 2013 Conference on Empirical Methods in Natural Language Processing}, pages 1631--1642, Seattle, Washington, USA. Association for Computational Linguistics.

\bibitem[{Stacey et~al.(2022)Stacey, Belinkov, and Rei}]{stacey2022supervising}
Joe Stacey, Yonatan Belinkov, and Marek Rei. 2022.
\newblock Supervising model attention with human explanations for robust natural language inference.
\newblock In \emph{Proceedings of the AAAI Conference on Artificial Intelligence}, volume~36, pages 11349--11357.

\bibitem[{Sundararajan et~al.(2017)Sundararajan, Taly, and Yan}]{sundararajan2017axiomatic}
Mukund Sundararajan, Ankur Taly, and Qiqi Yan. 2017.
\newblock Axiomatic attribution for deep networks.
\newblock In \emph{International conference on machine learning}, pages 3319--3328. PMLR.

\bibitem[{Tieleman and Hinton(2012)}]{tieleman2012rmsprop}
Tijmen Tieleman and Geoffrey Hinton. 2012.
\newblock \href {https://www.cs.toronto.edu/~tijmen/csc321/slides/lecture_slides_lec6.pdf} {Lecture 6.5---rmsprop: Divide the gradient by a running average of its recent magnitude}.
\newblock Lecture 6.5.
\newblock COURSERA: Neural networks for machine learning.

\bibitem[{Vaswani et~al.(2017)Vaswani, Shazeer, Parmar, Uszkoreit, Jones, Gomez, Kaiser, and Polosukhin}]{vaswani2017attention}
Ashish Vaswani, Noam Shazeer, Niki Parmar, Jakob Uszkoreit, Llion Jones, Aidan~N Gomez, {\L}ukasz Kaiser, and Illia Polosukhin. 2017.
\newblock Attention is all you need.
\newblock \emph{Advances in Neural Information Processing Systems}, 30.

\bibitem[{Wiegreffe and Pinter(2019)}]{wiegreffe2019attention}
Sarah Wiegreffe and Yuval Pinter. 2019.
\newblock Attention is not not explanation.
\newblock In \emph{Proceedings of the 2019 Conference on Empirical Methods in Natural Language Processing and the 9th International Joint Conference on Natural Language Processing (EMNLP-IJCNLP)}, pages 11--20.

\bibitem[{Wolf et~al.(2020)Wolf, Debut, Sanh, Chaumond, Delangue, Moi, Cistac, Rault, Louf, Funtowicz, Davison, Shleifer, von Platen, Ma, Jernite, Plu, Xu, Scao, Gugger, Drame, Lhoest, and Rush}]{wolf-etal-2020-transformers}
Thomas Wolf, Lysandre Debut, Victor Sanh, Julien Chaumond, Clement Delangue, Anthony Moi, Pierric Cistac, Tim Rault, Rémi Louf, Morgan Funtowicz, Joe Davison, Sam Shleifer, Patrick von Platen, Clara Ma, Yacine Jernite, Julien Plu, Canwen Xu, Teven~Le Scao, Sylvain Gugger, Mariama Drame, Quentin Lhoest, and Alexander~M. Rush. 2020.
\newblock \href {https://www.aclweb.org/anthology/2020.emnlp-demos.6} {Transformers: State-of-the-art natural language processing}.
\newblock In \emph{Proceedings of the 2020 Conference on Empirical Methods in Natural Language Processing: System Demonstrations}, pages 38--45, Online. Association for Computational Linguistics.

\bibitem[{Ying et~al.(2022)Ying, Hase, and Bansal}]{ying2022visfis}
Zhuofan Ying, Peter Hase, and Mohit Bansal. 2022.
\newblock Visfis: Visual feature importance supervision with right-for-the-right-reason objectives.
\newblock In \emph{Advances in Neural Information Processing Systems}.

\bibitem[{Zaheer et~al.(2020)Zaheer, Guruganesh, Dubey, Ainslie, Alberti, Ontanon, Pham, Ravula, Wang, Yang et~al.}]{zaheer2020big}
Manzil Zaheer, Guru Guruganesh, Kumar~Avinava Dubey, Joshua Ainslie, Chris Alberti, Santiago Ontanon, Philip Pham, Anirudh Ravula, Qifan Wang, Li~Yang, et~al. 2020.
\newblock Big bird: Transformers for longer sequences.
\newblock \emph{Advances in neural information processing systems}, 33:17283--17297.

\bibitem[{Zaidan and Eisner(2008)}]{zaidan-eisner-2008-modeling}
Omar Zaidan and Jason Eisner. 2008.
\newblock \href {https://aclanthology.org/D08-1004} {Modeling annotators: {A} generative approach to learning from annotator rationales}.
\newblock In \emph{Proceedings of the 2008 Conference on Empirical Methods in Natural Language Processing}, pages 31--40, Honolulu, Hawaii. Association for Computational Linguistics.

\bibitem[{Zhang et~al.(2015)Zhang, Zhao, and LeCun}]{zhang2015character}
Xiang Zhang, Junbo Zhao, and Yann LeCun. 2015.
\newblock Character-level convolutional networks for text classification.
\newblock \emph{Advances in neural information processing systems}, 28.

\end{thebibliography}

\clearpage
\appendix

\section{Attribution Techniques}
\label{sec_appendix:attribution_techniques}

We use a total of six attribution techniques in this work. They serve two main roles, acting as: \emph{(i)} \emph{guided attribution techniques}, when used as part of ER training to compute the explanation loss, $\mathcal{L}_{\mathrm{expl}}$; and \emph{(ii)} \emph{non-guided attribution techniques}, when used to evaluate the impact of ER on input attributions.
The attribution techniques we use are either \emph{gradient-based} or \emph{vector-based}. In the case of the latter, they differ on the components of the Transformer encoder architecture \citep{vaswani2017attention} that are used, and also on multiple design choices.
Finally, all techniques provide \emph{global} attributions, with the exception of attention, which is a \emph{local} attribution technique.

\paragraph{Attention.} Attention attributions correspond to the normalised top-layer attention weights. These are directly obtained from the Transformer model architecture. We use the values from the \textsc{[CLS]} token, averaged across heads.

\paragraph{Attention-Rollout.} Attention-Rollout \citep{abnar2020quantifying} attributions are obtained by recursively aggregating attention scores. As in the original implementation, we incorporate the residual connection by adding the identity matrix ($\mathbf{I}$) to the original attention matrix ($\mathbf{A}^l$), followed by a normalization step, resulting in $\mathbf{A}^l_R = 0.5 \mathbf{A}^l + 0.5\mathbf{I}$. We use the values from the \textsc{[CLS]} token, averaged across heads.

\paragraph{InputXGradient.} \textsc{InputXGradient} \citep{shrikumar2017learning} attributions are obtained by multiplying the input value by the corresponding gradient with respect to the output target label. We use the Captum package implementation \citep{kokhlikyan2020captum}.

\paragraph{ALTI.} ALTI \citep{ferrando2022measuring} is a vector-based attribution technique based on the attention block decomposition introduced in \citet{kobayashi-etal-2021-incorporating}. This decomposition factors in not only the attention weights, but also the value vectors, the output projection, as well as the first residual connection and layer normalization. The main difference lies in how the value of the contribution of a token at the layer level is computed --- ALTI takes into account the output vector of the attention block and uses the L1 norm instead of the L2 norm. Furthermore, ALTI then aggregates the local attributions over the full model using a rollout-like approach, resulting in global attributions.
We adapt the available code.\footnote{\href{https://github.com/mt-upc/transformer-contributions}{https://github.com/mt-upc/transformer-contributions}}

\paragraph{DecompX.} DecompX \citep{modarressi2023decompx}, similarly to ALTI, is a vector-based attribution technique based on the attention block decomposition by \citet{kobayashi-etal-2021-incorporating}. The main differences are twofold: first, DecompX incorporates all the Transformer encoder components, including the feed-forward networks. Second, instead of using a recursive approach like rollout to aggregate the local attributions, DecompX propagates `decomposed token representations' through the model.
We also report DecompX-Classifier, which includes the classification head, and whose output is signed.
We adapt the available code.\footnote{\href{https://github.com/mohsenfayyaz/DecompX}{https://github.com/mohsenfayyaz/DecompX}}

\section{Impact of Local vs Global Guided Attributions in ER}
\label{sec_appendix:global_attribution_supervision_signal}

We can use attention-rollout to show why we expect \emph{global} attributions to be more impactful than the \emph{local} counterparts when supervising a model with explanations.
Rollout was introduced in \citet{abnar2020quantifying}, and makes it possible to obtain global attributions by recursively aggregating vector-based local attributions, such as attention.
By defining attention-rollout recursively, $\mathbf a_l = \rho(\mathbf h^{l}, \mathbf h^{l-1}; \mathbf a_{l-1})$, where $\mathbf a_{l}$ corresponds to attention-rollout weights at layer $l$, and $\mathbf h^{l}$ to the hidden states at layer $l$, we can write its gradient. Namely, for the $l$th layer, we get:
\begin{equation}
\begin{aligned}
    \pdv{\mathbf a_l}{\theta} =&\pdv{\theta}\rho(\mathbf h^{l}, \mathbf h^{l-1}; \mathbf a_{l-1}) \\
   =& \pdv{\mathbf h^l}\rho(\mathbf h^{l}, \mathbf h^{l-1}; \mathbf a_{l-1}) \times \pdv{\mathbf h^l}{\theta} \\
   +& \pdv{\mathbf h^{l-1}}\rho(\mathbf h^{l}, \mathbf h^{l-1}; \mathbf a_{l-1})\times \pdv{\mathbf h^{l-1}}{\theta} \\
   +& \pdv{\mathbf a_{l-1}}\rho(\mathbf h^{l}, \mathbf h^{l-1}; \mathbf a_{l-1})\times \underbrace{\pdv{\mathbf a_{l-1}}{\theta}}_{\text{recursion}} . \\    
\end{aligned}
\label{eq:global_rollout}
\end{equation}

\begin{figure}
    \centering
    \begin{subfigure}[t]{0.45\columnwidth}
        \centering
        \includegraphics[height=1.8in]{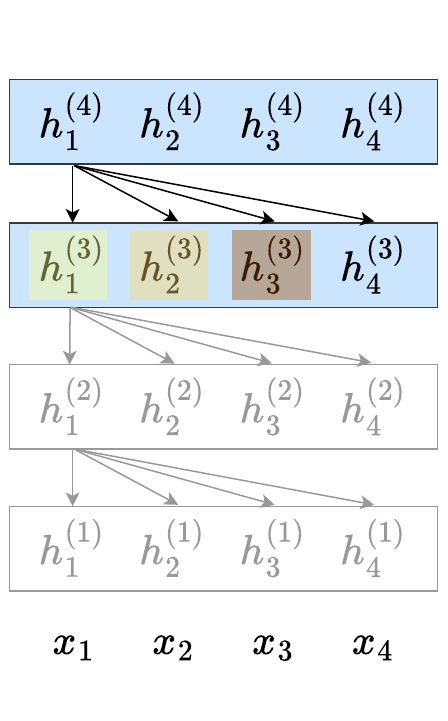}
        \caption{Local}
    \end{subfigure}%
    ~ 
    \begin{subfigure}[t]{0.45\columnwidth}
        \centering
        \includegraphics[height=1.8in]{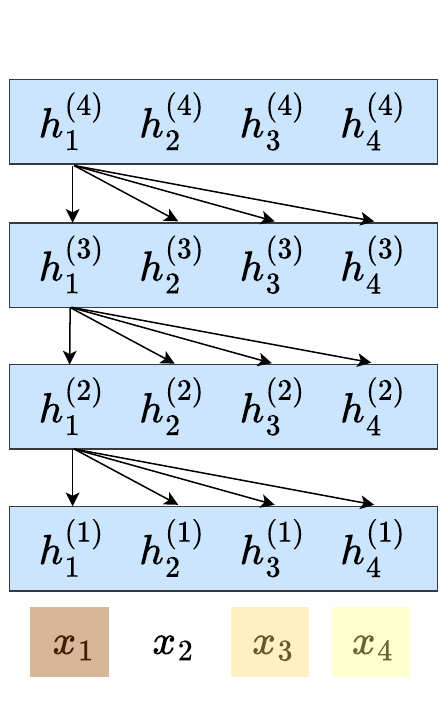}
        \caption{Global}
    \end{subfigure}
    \caption{Illustration of \emph{local} versus \emph{global} attribution techniques as tools for ER. Local attributes to the input of the top-layer. Global attributes to the input tokens using the full model, and potentially constraining the model more strongly to follow human annotations.}
\label{fig:global_vs_local}
\end{figure}

By inspecting Equation \ref{eq:global_rollout}, we can observe how the impact of guiding with a global vector-based technique goes beyond that of a local technique -- there we have $\mathbf a_l = \rho(\mathbf h^{l}, \mathbf h^{l-1})$, meaning that the last recursion term of the gradient that propagates to the layers below would not be part of the computation.
This difference seems to indicate that using a global attribution technique, such as attention-rollout, as $E$ in the explanation loss will more strongly limit the model's ability to condition on features other than the rationale tokens.
We show a visual interpretation of this difference in Figure \ref{fig:global_vs_local}.

\section{Attributions Techniques Faithfulness}

\begin{figure*}[t]
    \centering
    \includegraphics[width=0.8\linewidth]{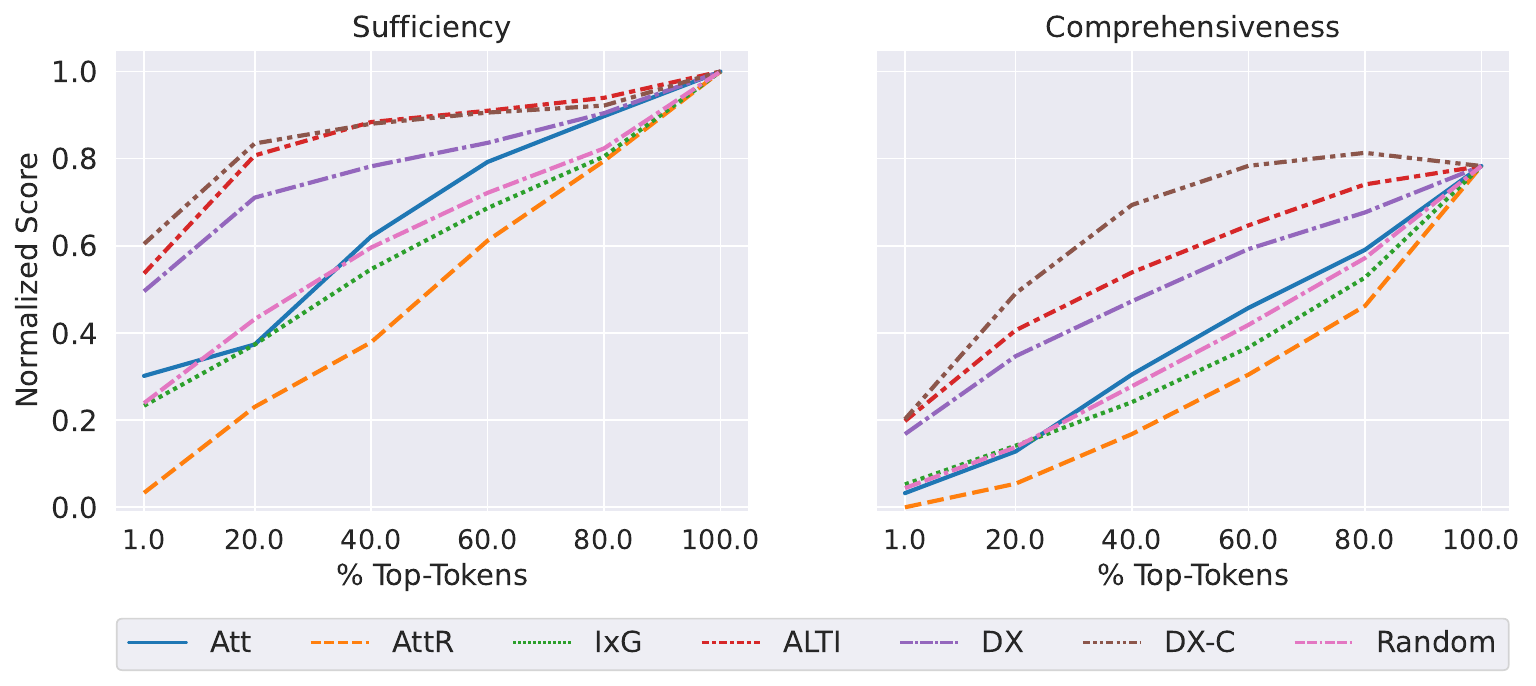}
\caption{Normalised sufficiency and comprehensiveness ($\uparrow$) scores for the top-x\% tokens according to a given attribution technique, for SST-Dev, using the \textsc{Baseline} approach.}
\label{fig:sa_suff_comp_baseline}
\end{figure*}

To further validate our choice of attribution techniques, we assess the faithfulness of the used input attributions by computing normalised sufficiency and comprehensiveness \citep{carton2020evaluating}. These are computed as follows:

\begin{equation}
    \mathrm{NullDiff}(x, \hat{y}) = \mathrm{max}(0, p(\hat{y}|x) - p(\hat{y}|x, 0))
\end{equation}

\begin{equation}
    \mathrm{NormSuff}(x, \hat{y}, \alpha) = \frac{\mathrm{Suff}(x, \hat{y}, \alpha) - \mathrm{Suff}(x, \hat{y}, 0)}{1 - \mathrm{Suff}(x, \hat{y}, \alpha)}
\end{equation}

\begin{equation}
    \mathrm{NormComp}(x, \hat{y}, \alpha) = \frac{\mathrm{Comp}(x, \hat{y}, \alpha)}{\mathrm{Comp}(x, \hat{y}, 1)},
\end{equation}

\noindent
with $\hat{y} = \mathrm{arg~max}~p(y|x)$.
Both normalised sufficiency and comprehensiveness, computed with the top 1\%, 20\%, 40\%, 60\%, 80\%, and 100\% most attributed tokens, can be seen in Figure \ref{fig:sa_suff_comp_baseline}.
As expected, both metrics improve the more top-attributed tokens, based on a given attribution technique, are used as input to the classifier.
Furthermore, we find DecompX, DecompX-Classifier, and ALTI to perform the best. This corroborates the results reported on their respective works.

\section{Plausibility Metrics}
\label{sec_appendix:plausibility_metrics}

We employ three metrics introduced in \citet{fomicheva2021eval4nlp,fomicheva2022translation} to measure how well the input attributions match the human annotated tokens.

\paragraph{AUC Score.} Computes the Area Under the Receiver Operating Characteristic Curve score, which considers multiple threshold values for the attributions with human annotations as the target label. This metric has been used in past works \cite{deyoung2020eraser, mathew2021hatexplain, fernandes2022scaffold, resck2024exploring}.

\paragraph{Average Precision.} As mentioned in \citet{fomicheva2021eval4nlp}, AUC scores might be optimistic when dealing with unbalanced data. That is relevant when working with rationales, where the annotated tokens might correspond to a small portion of the input. Thus, we report the average precision score, which summarizes the average-precision curve as a weighted average of the precision scores at the different thresholds,

\begin{equation}
    \textsc{AP} = \sum_n (R_n - R{n-1})~P_n,
\end{equation}

where $P_n$ and $R_n$ correspond to the precision and recall values at a given threshold.

\paragraph{Recall@k.} Measures the average recall for a specific number of tokens $k$. It is calculated as
\begin{equation}
    r@k = \frac{|\{x \in r_{k}(x)~:~x < k\}|}{N},
\end{equation}

where $N$ is the number of annotated tokens for $x$, and $r(.)$ is a function that retrieves the rank of all annotated tokens.

\section{Input Attributions Correlations}

\begin{figure*}[t!]
    \centering
    \begin{subfigure}[t]{0.99\linewidth}
        \centering
        \includegraphics[width=0.78\linewidth]{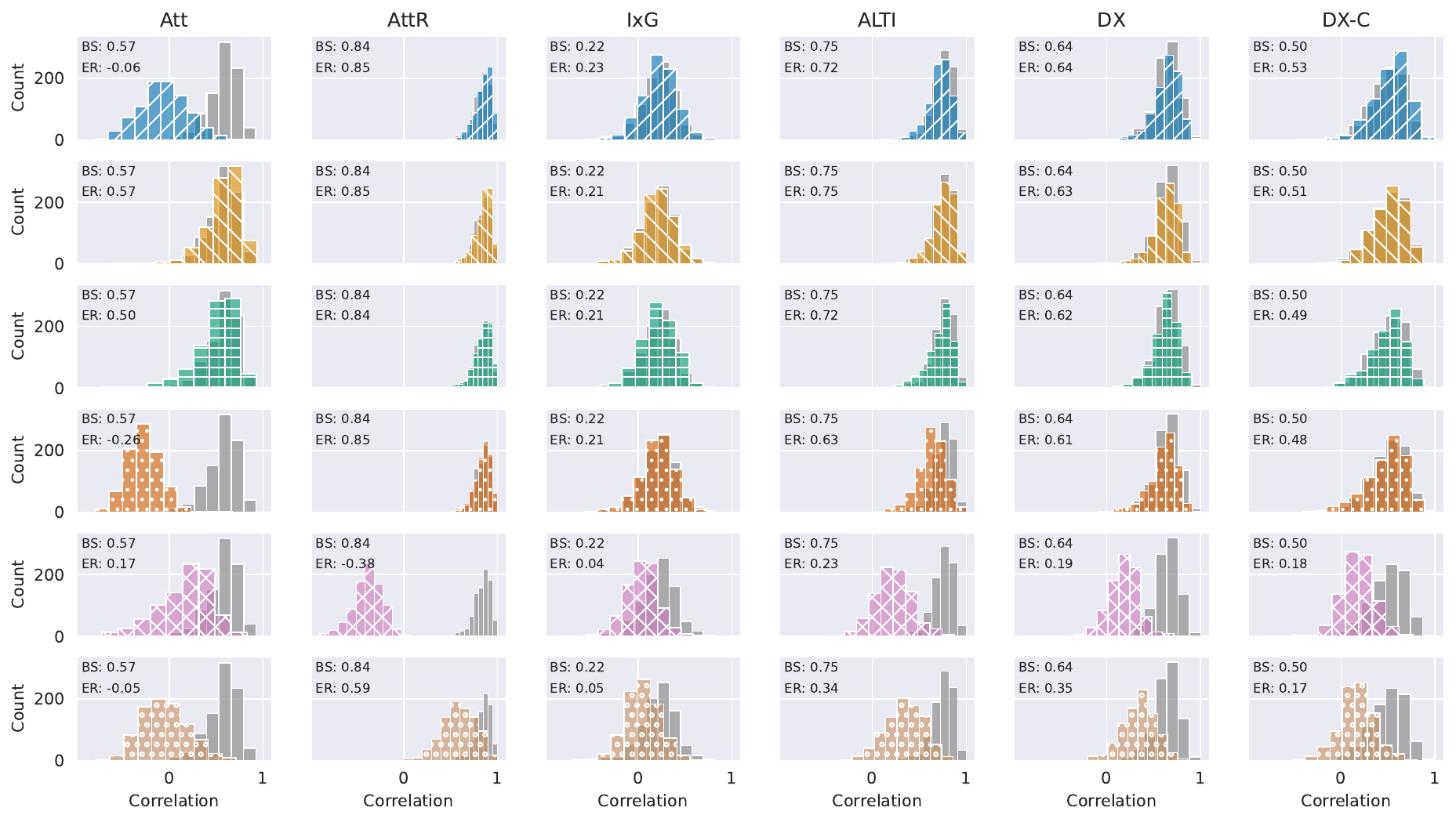}
        \caption{SST-Dev (ID)}
        \label{subfig:sa_attributions_correlations_approaches_all_sst-dev}
    \end{subfigure}%
    ~ \\
    \begin{subfigure}[t]{0.99\linewidth}
        \centering
        \includegraphics[width=0.38\linewidth]{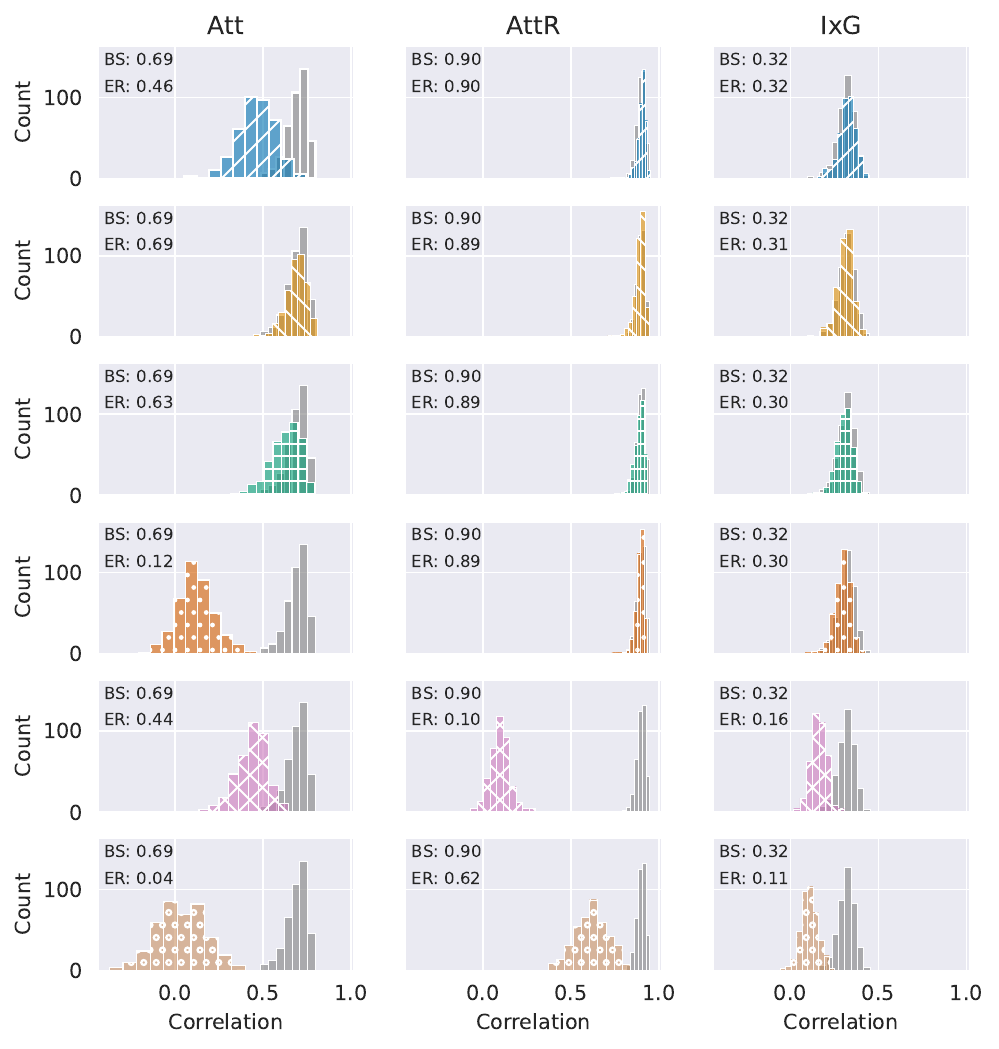}
        \caption{Movies (OOD)}
    \end{subfigure}%
    ~ \\
    \begin{subfigure}[t]{0.99\linewidth}
        \centering
        \includegraphics[width=0.78\linewidth]{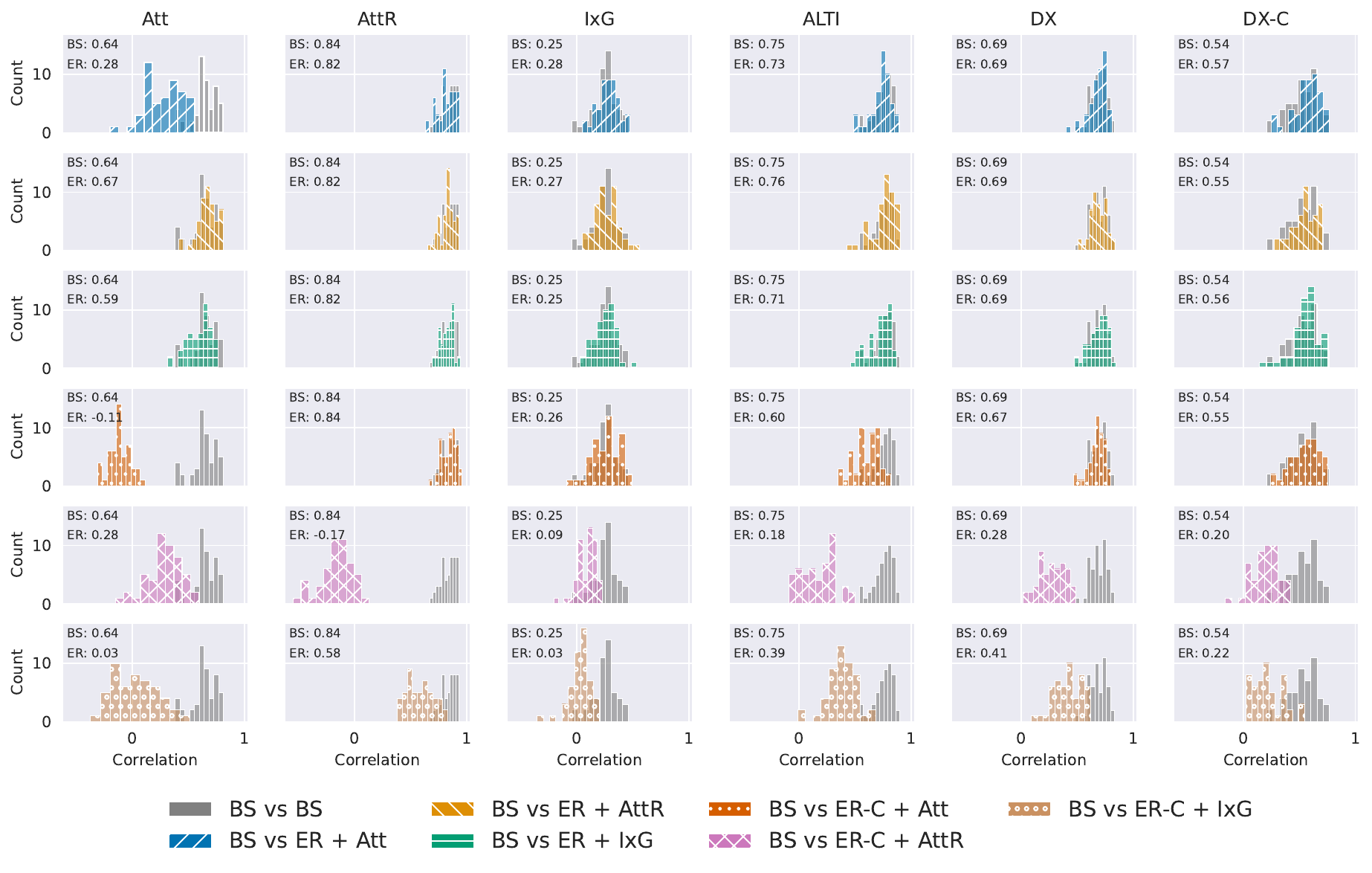}
        \caption{Yelp-50 (OOD)}
    \end{subfigure}
    \caption{Average Kendall Rank correlation between attribution techniques of the different approaches. The BS (Baseline) and ER (corresponding ER approach) text corresponds to the average correlation values.} 
\label{fig:sa_attributions_correlations_approaches_all}
\end{figure*}

As discussed in Section \ref{sec:results}, one of the tools we use to study the impact of ER on the input model attributions is to measure changes in the correlation of attribution scores for a fixed technique across approaches, \eg, how does attention correlate baseline vs baseline and baseline vs \textsc{ER+Att}.
Here, for each attribution technique, we iterate over all examples, sample two model versions, select the corresponding attributions for the \textsc{Baseline} and approach we want to assess, and then compute the respective Kendall rank correlation coefficient\footnote{\url{https://docs.scipy.org/doc/scipy/reference/generated/scipy.stats.kendalltau.html}}.
We show results for SST-Dev, Yelp-50, and Movies using all ER approaches and attribution techniques in Figure \ref{fig:sa_attributions_correlations_approaches_all}.
The conclusions are as discussed in Section \ref{sec:results}: ER techniques based on attention (\textsc{ER+Att} and \textsc{ER-C+Att}) impact only the attention attributions, while global constrained ER approaches (\textsc{ER-C+AttR} and \textsc{ER-C+IxG}) are able to impact all reported attribution techniques.

\begin{figure*}[t]
    \centering
    \begin{subfigure}[t]{0.98\linewidth}
        \centering
        \includegraphics[width=1\linewidth]{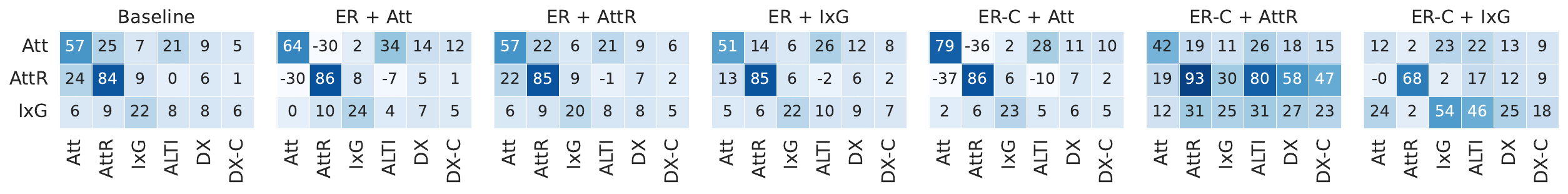}
        \caption{SST Dev (ID)}
    \end{subfigure}
    ~ 
    \begin{subfigure}[t]{0.98\linewidth}
        \centering
        \includegraphics[width=1\linewidth]{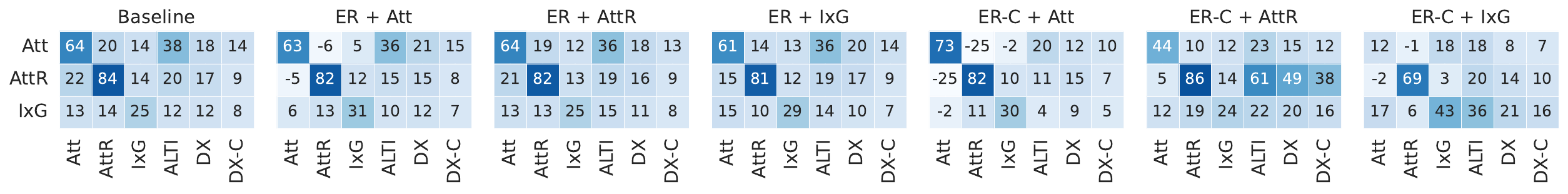}
        \caption{Yelp-50 (OOD)}
    \end{subfigure}
    ~ 
    \begin{subfigure}[t]{0.75\linewidth}
        \centering
        \includegraphics[width=1\linewidth]{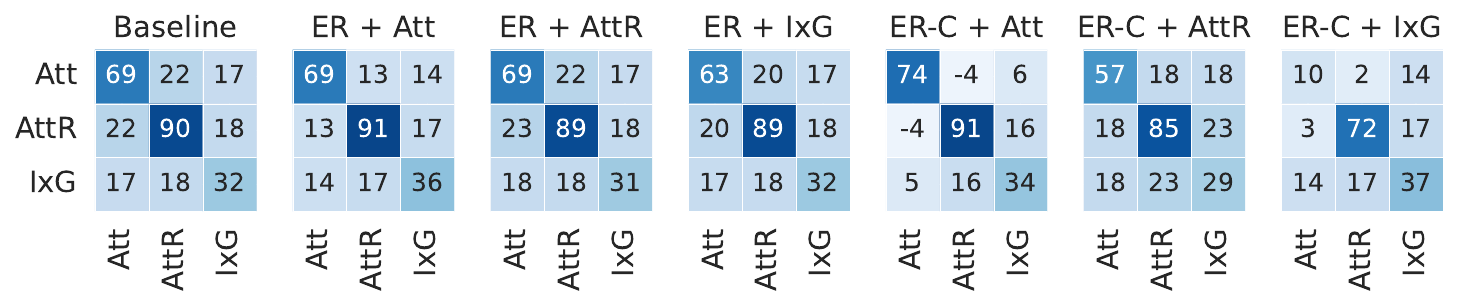}
        \caption{Movies (OOD)}
    \end{subfigure}
\caption{Average Kendall Rank correlation between attribution techniques for the different approaches.}
\label{fig:sa_attributions_correlations_techniques}
\end{figure*}

Alternatively, it is also possible to compare correlation coefficients for a pair of attribution techniques for a given choice of training objective (\eg, the correlation between attention and attention-rollout for baseline or \textsc{ER+Att}) and how those correlations change as we vary the training objective.
We compute average Kendall rank correlation values and report them in Figure \ref{fig:sa_attributions_correlations_techniques}.
The obtained results agree with our previous findings: \emph{(i)} attention as a guided attribution technique (\textsc{ER+Att} and \textsc{ER-C+Att}) impacts mostly only attention attributions; \emph{(ii)} global guided attribution techniques have little impact when used in joint ER (\textsc{ER+AttR} and \textsc{ER+IxG}); \emph{(iii)} only the constrained ER approaches with global guided attributions techniques are able to clearly impact most attributions; and \emph{(iv)} for OOD most patterns are the same, despite less noticeable.
Note how this correlation analysis is different from the previous. First, here we inspect averages (the number of data points makes it unfeasible to report histograms of correlation coefficients), which has the potential to be misleading.
Second, we correlate across attribution techniques instead of correlating the same attribution across approaches, which leads to observations that are more complex and difficult to interpret.


\section{Predicting OOD Performance}
\label{sec_appendix:predicting_ood_performance}

\begin{figure*}[t]
    \centering
    \begin{subfigure}[t]{0.98\linewidth}
        \centering
        \includegraphics[width=1\linewidth]{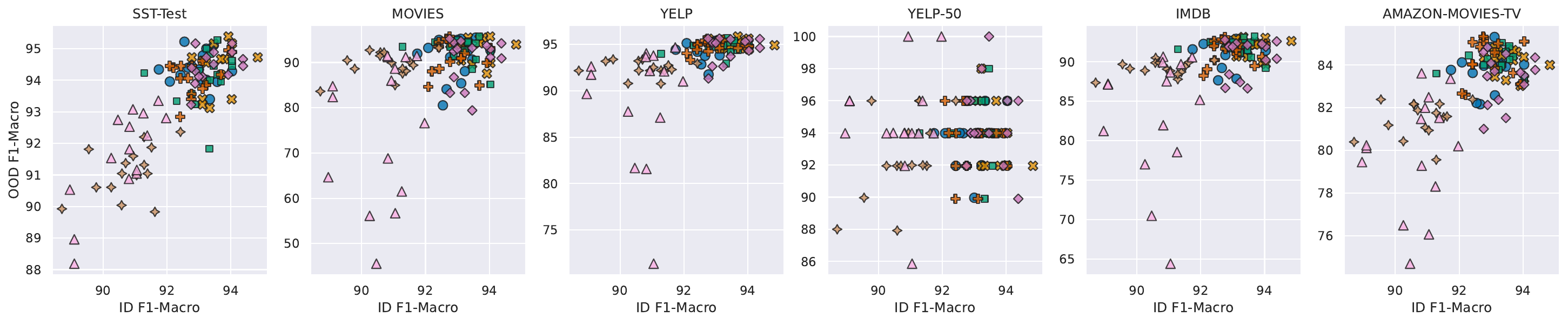}
        \caption{SST-Dev Classification as Predictor}
        \label{subfig:sa_ood_prediction_full_cls}
    \end{subfigure}
    ~ 
    \begin{subfigure}[t]{0.98\linewidth}
        \centering
        \includegraphics[width=1\linewidth]{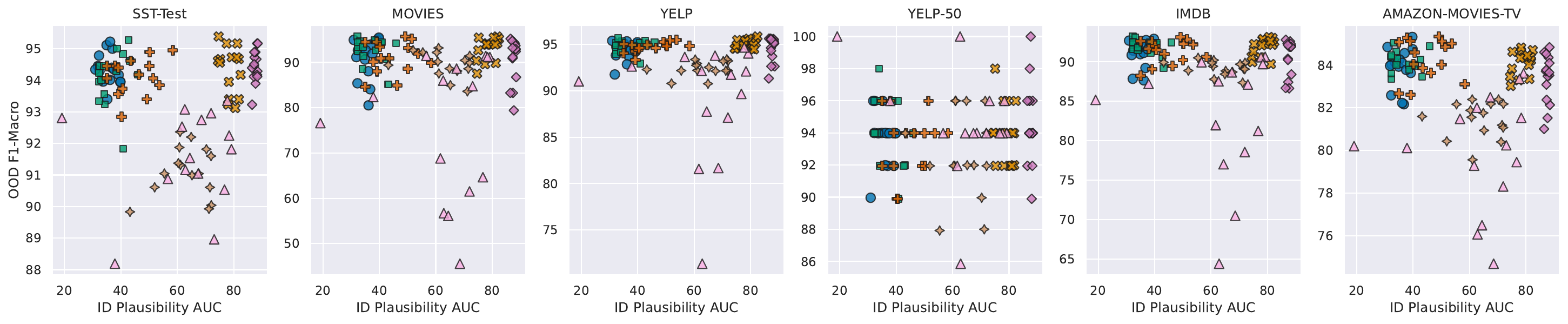}
        \caption{SST-Dev Plausibility as Predictor - AUC with Attention}
        \label{subfig:sa_ood_prediction_full_auc_attn}
    \end{subfigure}
    ~ 
    ~ 
    \begin{subfigure}[t]{0.98\linewidth}
        \centering
        \includegraphics[width=1\linewidth]{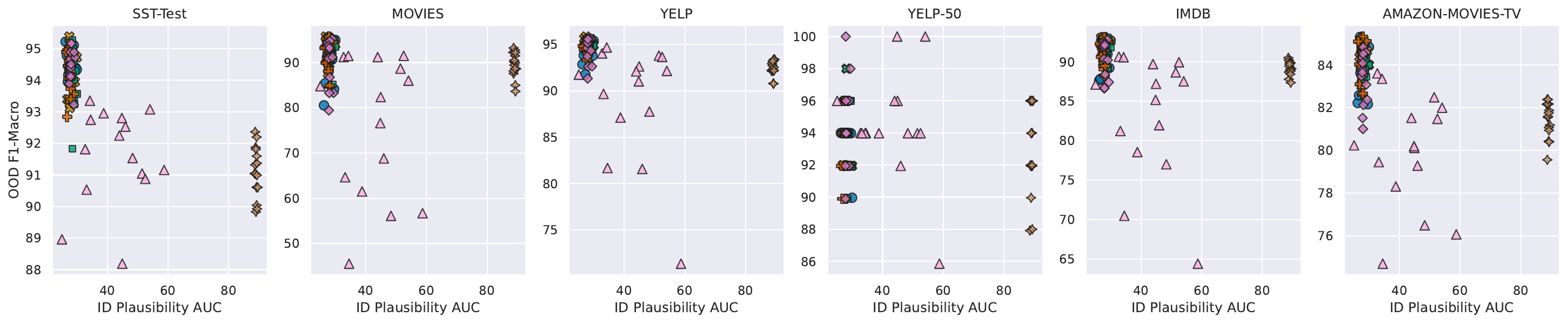}
        \caption{SST-Dev Plausibility as Predictor - AUC with Attention-Rollout}
        \label{subfig:sa_ood_prediction_full_auc_rollout}
    \end{subfigure}
    ~ 
    ~ 
    \begin{subfigure}[t]{0.98\linewidth}
        \centering
        \includegraphics[width=1\linewidth]{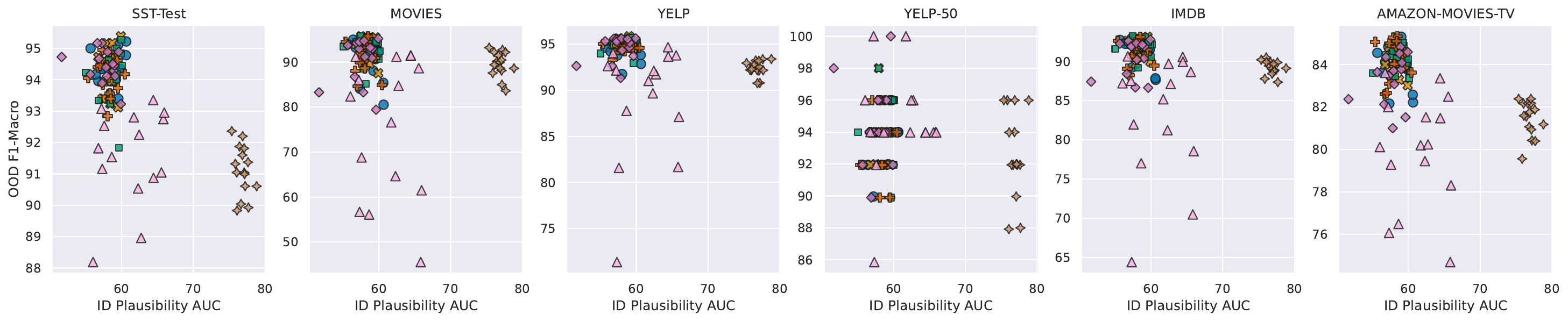}
        \caption{SST-Dev Plausibility as Predictor - AUC with DecompX-Classifier}
        \label{subfig:sa_ood_prediction_full_auc_dx-c}
    \end{subfigure}
    ~ 
    \begin{subfigure}[t]{0.98\linewidth}
        \centering
        \includegraphics[width=1\linewidth]{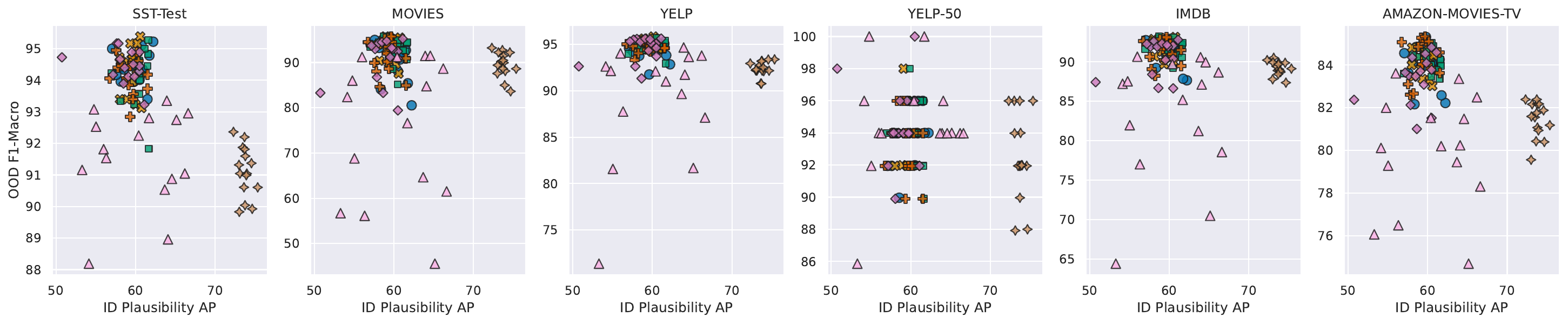}
        \caption{SST-Dev Plausibility as Predictor - Average Precision with DecompX-Classifier}
        \label{subfig:sa_ood_prediction_full_ap_dx-c}
    \end{subfigure}
    ~ 
    \begin{subfigure}[t]{0.98\linewidth}
        \centering
        \includegraphics[width=1\linewidth]{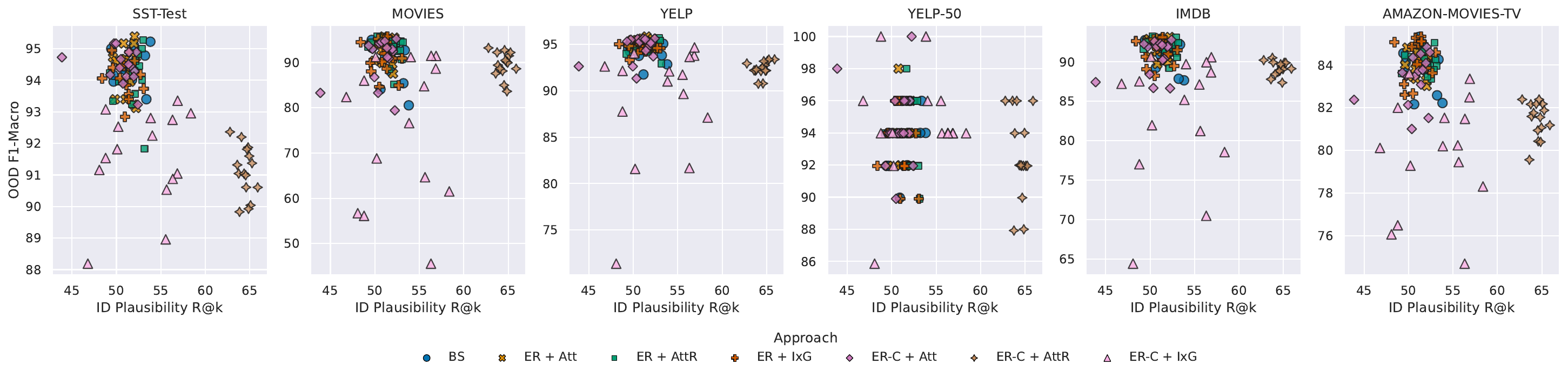}
        \caption{SST-Dev Plausibility as Predictor - Recall@k with DecompX-Classifier}
        \label{subfig:sa_ood_prediction_full_recall_dx-c}
    \end{subfigure}
    \caption{Scatter plot showing the relationship between in-domain (SST-Dev) measurements (F1-Macro classification scores and plausibility scores computed with different metrics and attribution techniques) for multiple datasets.}
\label{fig:sa_ood_prediction_full}
\end{figure*}

To complement the results discussed in Section \ref{subsec:predict_ood_performance}, we present findings on a broader range of OOD datasets and plausibility metrics criteria, also including joint \textsc{ER+Att} and \textsc{ER+AttR} results. Results are shown in Figure \ref{fig:sa_ood_prediction_full}. Similarly to Section \ref{subsec:predict_ood_performance}, no in-domain criteria exhibits correlation with OOD classification performance.

\section{Data}
\label{subsec_appendix:sst_data}

For in-domain and training data we use SST-2 \citep{socher-etal-2013-recursive}, following the heuristic algorithm proposed in \citet{carton2020evaluating} to obtain instance-level rationales\footnote{\url{https://github.com/BoulderDS/evaluating-human-rationales/blob/master/scripts/download_and_process_sst.py}}.
For out-of-domain we use: Amazon-Reviews\footnote{\url{https://huggingface.co/datasets/McAuley-Lab/Amazon-Reviews-2023}} \citep{hou2024bridging},
IMBD\footnote{\url{https://huggingface.co/datasets/stanfordnlp/imdb}} \citep{maas2011learningvectors},
Movies\footnote{\url{https://huggingface.co/datasets/eraser-benchmark/movie_rationales}} \citep{zaidan-eisner-2008-modeling, deyoung2020eraser},
and Yelp\footnote{\url{https://huggingface.co/datasets/fancyzhx/yelp_polarity}} \citep{zhang2015character}.
From these OOD datasets only Movies includes human annotated rationales.
Some OOD evaluation datasets have several thousand examples. In those cases we sample 5,000 examples. More details on the final data are described in Table \ref{tab:sa_data}. All data is in English.

\begin{table*}[t]
    \centering
    \scriptsize
    \begin{tabular}{c|ccc|cccc}
         \toprule
          & Train & Dev & Test & \% Positive & Avg. / Max Length & Source & Rationales \\
         \midrule
         \textsc{SST} \citep{socher-etal-2013-recursive} & 6,920 & 872 & 1,821 & $\approx$ 50~\% & $\approx 25$ / $\approx 65$& Movies & \checkmark \\
         \textsc{Movies} \citep{zaidan-eisner-2008-modeling,deyoung2020eraser} & - & - & 399$^{*}$ & 50.0~\% & 786 / 1024 & Movies & \checkmark \\
         \textsc{IMDB} \citep{maas2011learningvectors} & - & - & 5,000 & 50.7~\% & 288 / 1024 & Movies & \newcrossmark \\
         \textsc{Amazon (Movies and TV)} \citep{hou2024bridging} & - & - & 5,000 & 77.5~\% & 107 / 1024 & Products & \newcrossmark \\
         \textsc{Yelp} \citep{zhang2015character} & - & - & 5,000 & 50.6~\% & 168 / 1024 & Businesses & \newcrossmark \\
         \textsc{Yelp-50}$^{**}$ & - & - & 50 & 50.0~\% & 84 / 151 & Businesses & \checkmark \\
         \bottomrule
    \end{tabular}
    \caption{Sentiment analysis datasets, including percentage of examples labeled as positive, average and maximum length of inputs (after truncating), source of the data, and whether rationales are provided. $^{*}$ -- This includes both dev and test examples. $^{**}$ -- This corresponds to the subset of annotated data described in Section \ref{sec_appendix:extra_ood_annotations}.}
    \label{tab:sa_data}
\end{table*}

\subsection{Extra OOD Annotated Data}
\label{sec_appendix:extra_ood_annotations}

We conduct our OOD analysis on the Movies dataset, an available sentiment analysis dataset with human rationale annotations.
To further support our claims with respect to increased robustness to OOD conditions, we annotated a split of 50 examples of the Yelp dataset.
We first filter out examples that will result in more than 150 subtokens when using the tokenizer from the pre-trained encoder model we use during training (so that we can compute attributions with all available metrics, in particular ALTI \citep{ferrando2022measuring} and DecompX \citep{modarressi2023decompx}, which we find to result in prohibitive runtime or OOM errors with long sequences).
Then, we sample 55 examples at random from those, tokenize them using the English tokenizer from spaCy \citep{spacy2020}, and manually annotate the first five, selecting all sequences of text that offer evidence for the gold label.
These are used as few-shot examples for a prompt to \textsc{Meta-Llama-3-8B-Instruct}\footnote{\url{https://huggingface.co/meta-llama/Meta-Llama-3-8B-Instruct}} \citep{llama3modelcard} that outputs automatic annotations for the remaining 50 examples.
Finally, we post-edit the automatic annotations whenever necessary.

Average model classification performance for this split of data is shown in Table \ref{tab:sa_yelp_50_ood}.

\begin{table}
    \centering
    \small
    \begin{tabular}{lc}
        \toprule
         & Average F1-Macro \\
         \midrule
        \textsc{Baseline} & 93.71 $\pm$ 1.62 \\
        \midrule
        \textsc{ER+Att} & 93.98 $\pm$ 1.80 \\
        \textsc{ER+AttR} & 94.24 $\pm$ 2.07 \\
        \textsc{ER+IxG} & 93.30 $\pm$ 1.91 \\
        \midrule
        \textsc{ER-C+Att} & 94.64 $\pm$ 2.41 \\
        \textsc{ER-C+AttR} & 92.64 $\pm$ 2.61 \\
        \textsc{ER-C+IxG} & 94.51 $\pm$ 3.17 \\
        \bottomrule
    \end{tabular}
    \caption{F1-Macro scores ($\uparrow$) for \textsc{Yelp-50}.}
    \label{tab:sa_yelp_50_ood}
\end{table}


\section{Hyperparameters and Training}
\label{sec_appendix:training}

\paragraph{Optimizer and Scheduler.} We use AdamW \citep{loshchilov2018decoupled} as our optimizer, with $\beta$ values set to (0.9, 0.98) and the weight decay coefficient set to 0.
For the learning rate we use a linear scheduler, with 10\% of training steps as warm-up.
For constrained optimisation parameters we use RMSprop \citep{tieleman2012rmsprop}.

\paragraph{Hyperparameter Selection.}

We choose the combination of learning rate and maximum number of training epochs (since we use a linear scheduler with 10\% of training steps as warm-up) with the lowest cross-entropy loss over three runs. For the explanation regularised approach we explore the same hyperparameters, plus the $\lambda$ weight. We select based on cross-entropy loss so that different values of $\lambda$ can be directly compared.

For the constrained approach, we first train a model (3 seeds) using $\mathcal{L} = \mathcal{L}_{\mathrm{expl}}$ and use the average minimum explanation train and validation explanation losses to guide our choice of bounds $b_{\mathrm{train}}$ and $b_{\mathrm{val}}$.
For $b_{\mathrm{val}}$ we use the average of the minimum validation loss.
For $b_{\mathrm{train}}$ we use a value close to 1.5 times the average of the minimum training loss.
Then, we choose the combination of learning rate and constrained optimizer learning rate that minimizes the average cross entropy loss.

For the baseline we explore learning rate $\in \{2, 3, 5 \times 10^{-5}\}$ and maximum number of epochs $\in \{15,25\}$. For the joint approach we explore the same space as the baseline model and  $\lambda \in \{0.6, 1.0, 1.4\}$, following a set of choices aligned with previous work on ER for OOD robustness.
For the constrained approach we explore learning rate $\in \{2, 3, 5 \times 10^{-5}\}$ and constrained learning rate $\in \{1 \times 10^{-1}, 5 \times 10^{-2}\}$.
All experiments use a batch size of 32.
Final choices can be seen in Table \ref{tab:sa_hparams}.
The experiments with multiple lambdas uses the same choices as the baseline model.

\begin{table}[t]
\centering
\footnotesize
\begin{tabular}{l|l}
\toprule
\textbf{Baseline} & \textbf{Value} \\
\midrule
Learning Rate & $3 \times 10^{-5}$\\
Train Epochs & 25 \\
\midrule
\textbf{Joint Attention} & \textbf{Value} \\
\midrule
Learning Rate & $2 \times 10^{-5}$\\
Train Epochs & 25 \\
$\mathbf{\lambda}$ & 1.0 \\
\midrule
\textbf{Joint Rollout} & \textbf{Value} \\
\midrule
Learning Rate & $3 \times 10^{-5}$ \\
Train Epochs & 15 \\
$\mathbf{\lambda}$ & 1.0 \\
\midrule
\textbf{Joint IxG} & \textbf{Value} \\
\midrule
Learning Rate & $2 \times 10^{-5}$ \\
Train Epochs & 25 \\
$\mathbf{\lambda}$ & 1.0 \\
\midrule
\textbf{Constrained Attention} & \textbf{Value} \\
\midrule
Learning Rate & $2 \times 10^{-5}$ \\
Train Epochs & 25 \\
Constrained Learning Rate & $5 \times 10^{-2}$ \\
$b_{\mathrm{train}}$ & 0.035 \\
$b_{\mathrm{val}}$ & 0.031 \\
\midrule
\textbf{Constrained Rollout} & \textbf{Value} \\
\midrule
Learning Rate & $3 \times 10^{-5}$ \\
Train Epochs & 25 \\
Constrained Learning Rate & $1 \times 10^{-1}$ \\
$b_{\mathrm{train}}$ & 0.030 \\
$b_{\mathrm{val}}$ & 0.031 \\
\midrule
\textbf{Constrained IxG} & \textbf{Value} \\
\midrule
Learning Rate & $2 \times 10^{-5}$ \\
Train Epochs & 25 \\
Constrained Learning Rate & $1 \times 10^{-1}$ \\
$b_{\mathrm{train}}$ & 0.35 \\
$b_{\mathrm{val}}$ & 0.35 \\
\bottomrule
\end{tabular}
\caption{Hyperparameters choices for all approaches.}
\label{tab:sa_hparams}
\end{table}

\paragraph{Training.}

During training we choose a model checkpoint based on average validation loss for the baseline and the \emph{joint} explanation regularisation approach.
This follows \citet{joshi-etal-2022-er} and ensures that the explanation loss of the ER approach directly influences model selection.

For constrained optimization, we choose the model checkpoint with the lowest validation cross-entropy loss, provided that the validation explanation loss is $\mathcal{L}_{\mathrm{expl}_{\mathrm{val}}} < 1.1 \times b_{\mathrm{val}}$.
This ensures that we compare model checkpoints where the guided attribution technique is learning to predict the annotated rationales, according to the defined bound.
Selecting a checkpoint based on the same criteria would have not been possible for the joint approach, as none of the runs converges to an explanation loss value that meets the defined validation bound.

Unless mentioned otherwise, we report results over 15 seeds.\footnote{We found \textsc{ER-C+IxG} unstable to train, with only around one in three seeds minimising the CE loss while also minimising the explanation loss. Thus, it required training more models. In practice, we still report results over 15 seeds.}
All our experiments are developed using a single Nvidia A100 40GB GPU, and implemented with PyTorch \citep{paszke2019pytorch} and PyTorch Lightning \citep{Falcon_PyTorch_Lightning_2019}.

\paragraph{Attribution Techniques.}

Following \citet{joshi-etal-2022-er}, we scale the attributions $E (C_\theta, x)$ by 100 when computing the explanation loss and re-normalize them with softmax. We take it as part of the approaches that use attention as the guided attribution technique, \textsc{ER+Att} and \textsc{ER-C+Att}.
For \textsc{InputXGradient} we keep all default choices of the Captum package. The output attributions are aggregated into token-level attributions via sum, and normalised with the L2 norm.
For \textsc{DecompX}/\textsc{DecompX-C}\footnote{\url{https://github.com/mohsenfayyaz/DecompX}}, and \textsc{ALTI}\footnote{\url{https://github.com/mt-upc/transformer-contributions}} we keep all default choices part of the original work.
For attributions techniques that output `signed' attribution scores, \ie, \textsc{IxG} and \textsc{DX-C}, we take the absolute value of the attribution score.

\paragraph{Classifier.}

Following \citet{joshi-etal-2022-er}, we use a pre-trained Transformer encoder model, \textsc{google/bigbird-roberta-base}\footnote{\url{https://huggingface.co/google/bigbird-roberta-base}} \citep{zaheer2020big}, followed by a linear layer that uses as input the top-layer representation of the \textsc{[CLS]} token, with \textsc{tanh} as the non-linearity, and no classifier dropout.

\newpage

\begin{figure*}
    \centering
    \includegraphics[width=1.0\linewidth]{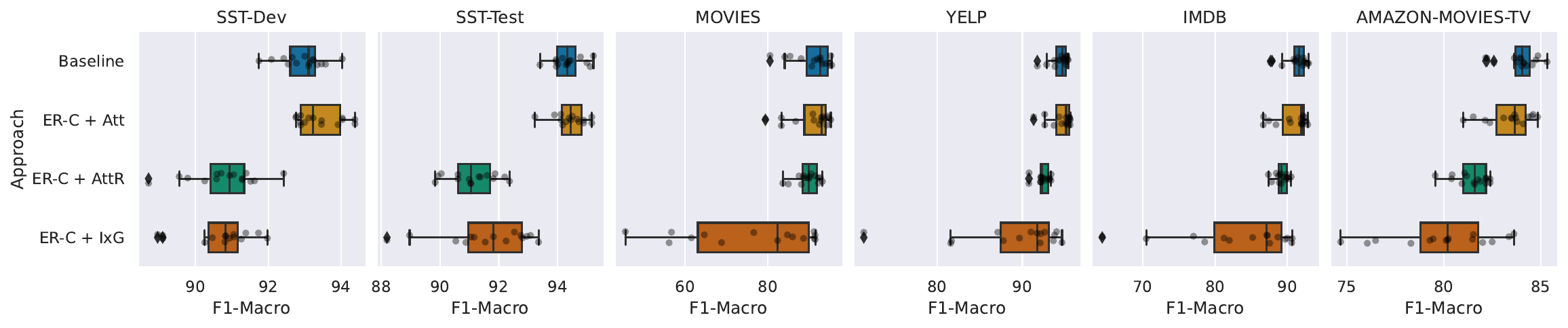}
    \caption{F1-Macro scores ($\uparrow$). \textsc{ER-C + Att} uses attention as the guided attribution technique, \textsc{ER-C + AttR} uses attention-rollout, and \textsc{ER-C + IxG} uses \textsc{InputXGradient}. Results correspond to 15 seeds.}
    \label{fig:sa_results_constrained}
\end{figure*}
\begin{figure*}[t]
    \centering
        \centering
        \includegraphics[width=0.45\linewidth]{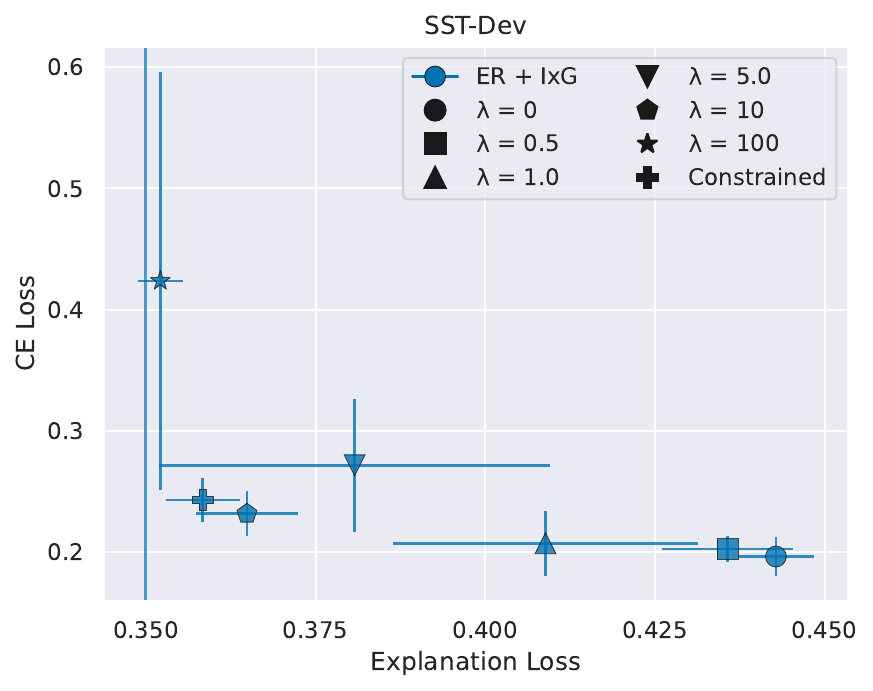}
    \caption{Cross-Entropy vs Explanation Loss for SST-Dev, using \textsc{ER+IxG}. The vertical line shows the validation explanation-loss bound. Each point corresponds to the average of 5 runs, and the error bars to the standard deviation for each loss. $\lambda$ ranges from 0 (No-ER) to 100.}
\label{fig:sa_multiple-lambdas_ixg}
\end{figure*}

\begin{figure*}[t!]
    \centering
    \begin{subfigure}[t]{1.0\linewidth}
        \centering
        \includegraphics[height=2.4in]{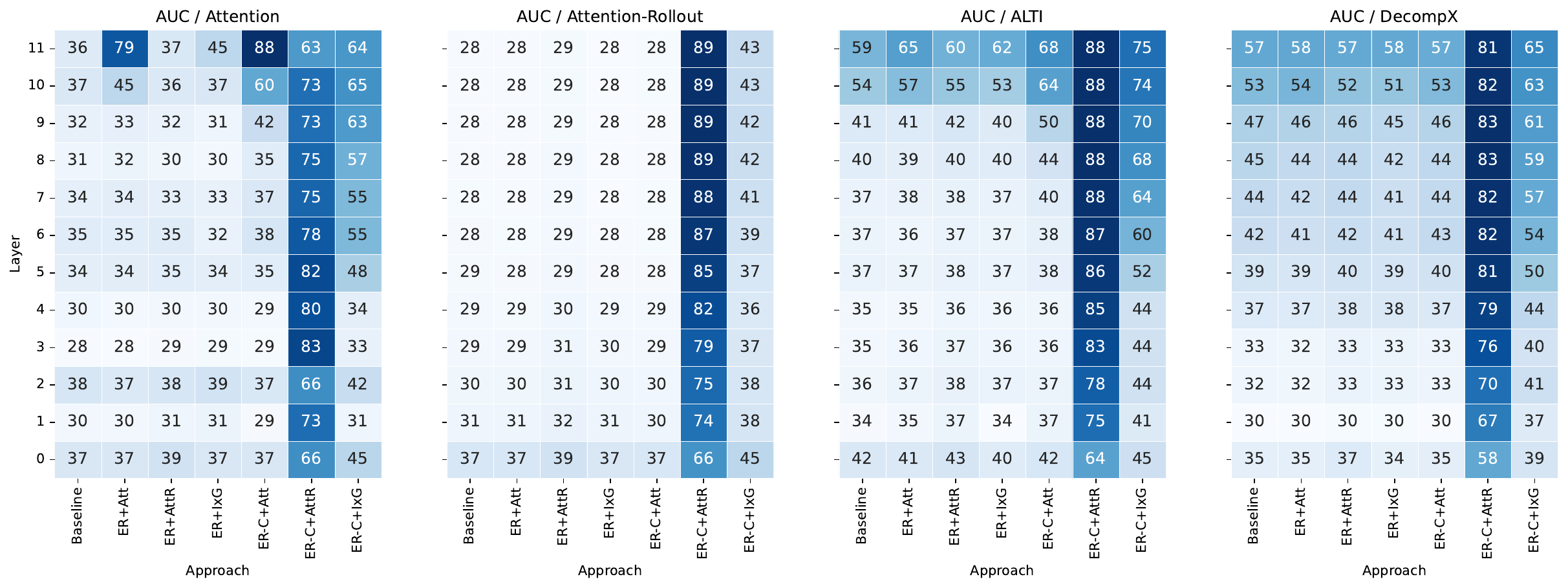}
        \caption{SST-Dev (ID)}
    \end{subfigure}%
    ~ \\
    \begin{subfigure}[t]{1.0\linewidth}
        \centering
        \includegraphics[height=2.4in]{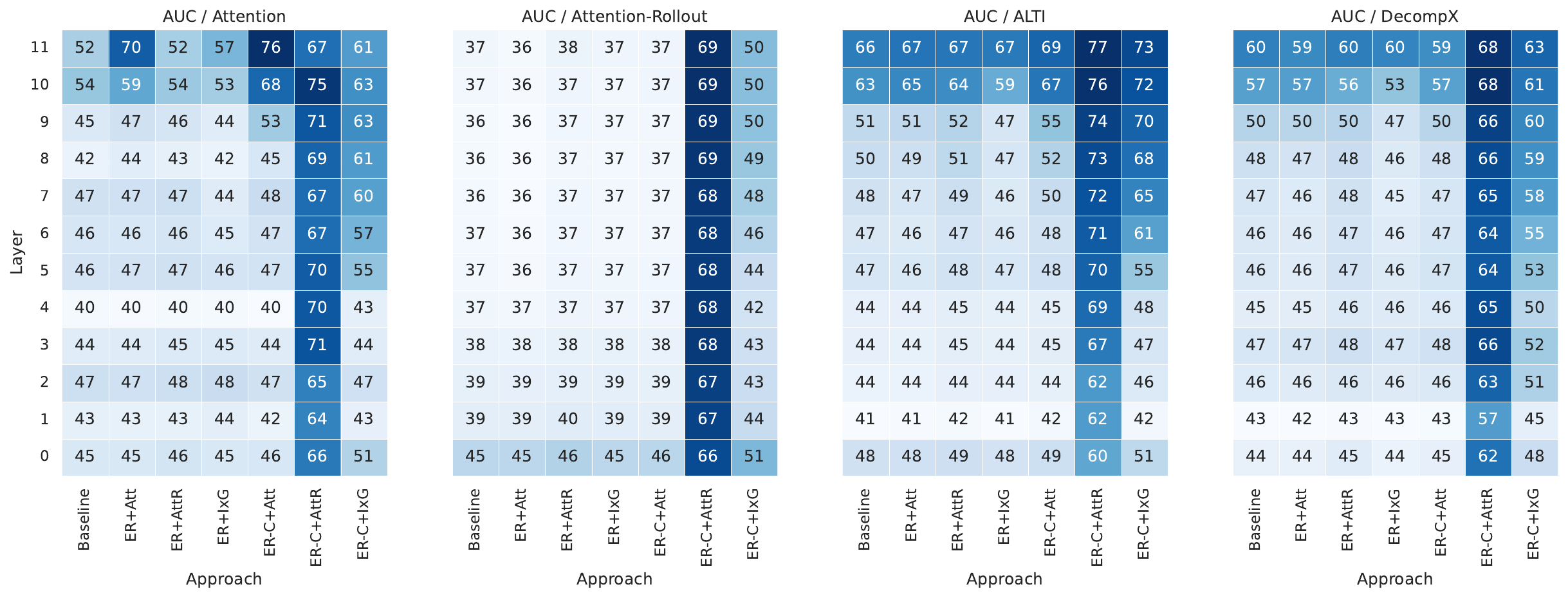}
        \caption{Yelp-50 (OOD)}
    \end{subfigure}%
    ~ \\
    \begin{subfigure}[t]{1.0\linewidth}
        \centering
        \includegraphics[height=2.4in]{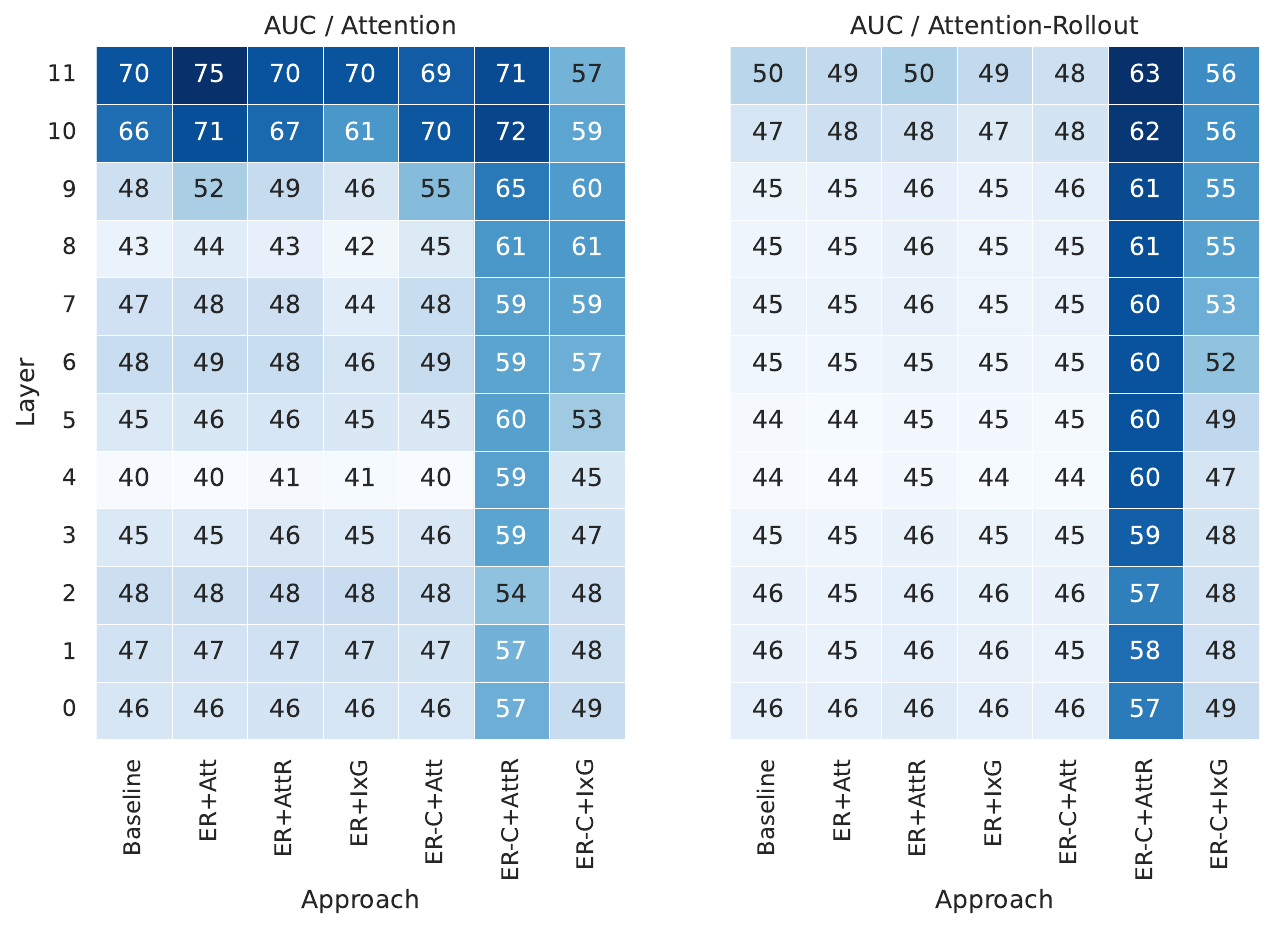}
        \caption{Movies (OOD)}
    \end{subfigure}
    \caption{Average AUC plausibility scores per-layer ($\uparrow$). We report only techniques that output per-layer attributions.}
\label{fig:appendix_auc_per_layer_all}
\end{figure*}

\begin{table*}
    \addtolength{\tabcolsep}{-1.5pt}
    \centering
    \tiny
    \rotatebox{90}{
        \begin{minipage}{\textheight}
        \centering
        \begin{tabular}{lcccccc|cccccc|cccccc}
            \toprule
            & \multicolumn{6}{c}{AUC} & \multicolumn{6}{c}{AP} & \multicolumn{6}{c}{R@k} \\
            \midrule
            & \textsc{Att} & \textsc{AttR} & \textsc{IxG} & \textsc{ALTI} & \textsc{DX-C} & \textsc{DX} &
            \textsc{Att} & \textsc{AttR} & \textsc{IxG} & \textsc{ALTI} & \textsc{DX-C} & \textsc{DX} &
            \textsc{Att} & \textsc{AttR} & \textsc{IxG} & \textsc{ALTI} & \textsc{DX-C} & \textsc{DX} \\
            \midrule
            \multicolumn{19}{c}{\emph{SST-Dev (ID)}} \\
            \midrule
            \textbf{Baseline} & 35.6 $\pm$ 2.9 & 27.9 $\pm$ 1.1 & 50.7 $\pm$ 1.5 & 59.0 $\pm$ 2.8 & 58.5 $\pm$ 1.6 & 57.5 $\pm$ 1.7 & 39.8 $\pm$ 2.0 & 35.6 $\pm$ 0.4 & 50.7 $\pm$ 1.0 & 61.2 $\pm$ 2.2 & 59.8 $\pm$ 1.6 & 58.3 $\pm$ 1.7 & 32.0 $\pm$ 1.9 & 28.0 $\pm$ 0.6 & 42.7 $\pm$ 0.9 & 52.3 $\pm$ 2.2 & 51.4 $\pm$ 1.3 & 50.4 $\pm$ 1.4 \\
            \textbf{ER + Att} & 78.6 $\pm$ 2.8 & 27.9 $\pm$ 0.8 & 50.6 $\pm$ 1.8 & 64.5 $\pm$ 2.4 & 58.4 $\pm$ 0.9 & 57.8 $\pm$ 1.6 & 76.7 $\pm$ 2.2 & 35.7 $\pm$ 0.4 & 50.5 $\pm$ 1.5 & 64.1 $\pm$ 1.3 & 59.5 $\pm$ 0.9 & 58.0 $\pm$ 1.4 & 67.9 $\pm$ 2.3 & 28.3 $\pm$ 0.5 & 42.5 $\pm$ 1.2 & 55.5 $\pm$ 1.4 & 51.2 $\pm$ 0.9 & 50.3 $\pm$ 1.3 \\
            \textbf{ER + AttR} & 37.0 $\pm$ 4.7 & 28.7 $\pm$ 0.6 & 50.8 $\pm$ 2.1 & 60.2 $\pm$ 3.0 & 58.6 $\pm$ 1.4 & 57.4 $\pm$ 2.0 & 40.8 $\pm$ 3.1 & 35.9 $\pm$ 0.3 & 50.5 $\pm$ 1.9 & 61.9 $\pm$ 2.1 & 60.2 $\pm$ 1.4 & 58.6 $\pm$ 2.0 & 32.9 $\pm$ 3.6 & 28.4 $\pm$ 0.4 & 42.8 $\pm$ 1.4 & 53.3 $\pm$ 2.1 & 51.8 $\pm$ 1.2 & 50.8 $\pm$ 1.6 \\
            \textbf{ER + IxG} & 44.6 $\pm$ 7.2 & 27.6 $\pm$ 0.7 & 53.8 $\pm$ 4.7 & 61.7 $\pm$ 4.6 & 58.0 $\pm$ 1.2 & 57.6 $\pm$ 1.7 & 47.0 $\pm$ 5.5 & 35.5 $\pm$ 0.2 & 53.7 $\pm$ 4.3 & 62.6 $\pm$ 3.4 & 59.3 $\pm$ 1.3 & 57.8 $\pm$ 1.5 & 39.3 $\pm$ 5.3 & 27.9 $\pm$ 0.4 & 45.7 $\pm$ 4.0 & 53.8 $\pm$ 3.4 & 51.0 $\pm$ 1.2 & 50.0 $\pm$ 1.4 \\
            \textbf{ER-C + Att} & 87.6 $\pm$ 0.7 & 28.2 $\pm$ 0.5 & 51.5 $\pm$ 1.4 & 68.3 $\pm$ 3.6 & 57.6 $\pm$ 2.0 & 56.8 $\pm$ 1.5 & 84.8 $\pm$ 0.5 & 35.7 $\pm$ 0.2 & 51.2 $\pm$ 1.2 & 65.8 $\pm$ 2.1 & 58.6 $\pm$ 2.5 & 56.6 $\pm$ 1.6 & 76.3 $\pm$ 0.6 & 28.3 $\pm$ 0.4 & 43.2 $\pm$ 1.0 & 57.9 $\pm$ 2.2 & 50.6 $\pm$ 2.1 & 49.2 $\pm$ 1.4 \\
            \textbf{ER-C + AttR} & 63.2 $\pm$ 8.4 & 89.1 $\pm$ 0.3 & 67.6 $\pm$ 1.5 & 87.9 $\pm$ 0.6 & 76.9 $\pm$ 0.9 & 81.3 $\pm$ 0.6 & 63.9 $\pm$ 7.7 & 85.6 $\pm$ 0.5 & 61.8 $\pm$ 1.1 & 84.9 $\pm$ 0.6 & 73.7 $\pm$ 0.7 & 77.0 $\pm$ 0.5 & 55.1 $\pm$ 7.6 & 77.1 $\pm$ 0.6 & 53.8 $\pm$ 1.1 & 76.5 $\pm$ 0.7 & 64.5 $\pm$ 0.8 & 67.9 $\pm$ 0.6 \\
            \textbf{ER-C + IxG} & 63.9 $\pm$ 16 & 42.9 $\pm$ 9.5 & 81.5 $\pm$ 2.8 & 74.8 $\pm$ 3.1 & 61.3 $\pm$ 3.6 & 64.8 $\pm$ 4.1 & 64.2 $\pm$ 14 & 42.4 $\pm$ 4.9 & 83.2 $\pm$ 2.5 & 75.3 $\pm$ 3.0 & 60.4 $\pm$ 4.9 & 62.1 $\pm$ 5.6 & 55.5 $\pm$ 14 & 37.1 $\pm$ 6.4 & 74.0 $\pm$ 2.7 & 67.3 $\pm$ 2.7 & 53.1 $\pm$ 3.9 & 55.7 $\pm$ 4.6 \\
            \midrule
            $\mathcal L_{\text{expl}}$ (A) & 89.0 $\pm$ 0.4 & 29.6 $\pm$ 0.5 & 53.2 $\pm$ 1.6 & 78.2 $\pm$ 1.2 & 52.0 $\pm$ 0.7 & 53.0 $\pm$ 0.8 & 87.4 $\pm$ 0.3 & 36.0 $\pm$ 0.1 & 51.8 $\pm$ 0.9 & 73.8 $\pm$ 0.2 & 50.9 $\pm$ 1.0 & 50.0 $\pm$ 1.6 & 79.2 $\pm$ 0.3 & 28.6 $\pm$ 0.0 & 43.6 $\pm$ 0.6 & 65.0 $\pm$ 0.2 & 44.0 $\pm$ 0.5 & 43.6 $\pm$ 1.8 \\
            $\mathcal L_{\text{expl}}$ (R) & 81.1 $\pm$ 2.4 & 89.4 $\pm$ 0.2 & 72.9 $\pm$ 1.4 & 88.2 $\pm$ 0.5 & 77.1 $\pm$ 2.5 & 83.3 $\pm$ 1.6 & 76.3 $\pm$ 1.8 & 86.7 $\pm$ 0.4 & 65.4 $\pm$ 0.6 & 84.0 $\pm$ 0.8 & 72.3 $\pm$ 1.6 & 77.6 $\pm$ 1.4 & 69.6 $\pm$ 2.1 & 78.6 $\pm$ 0.4 & 57.5 $\pm$ 0.6 & 76.3 $\pm$ 0.9 & 63.5 $\pm$ 1.6 & 69.4 $\pm$ 1.7 \\
            $\mathcal L_{\text{expl}}$ (IxG) & 69.2 $\pm$ 6.3 & 57.4 $\pm$ 13 & 84.3 $\pm$ 0.5 & 81.0 $\pm$ 1.0 & 65.2 $\pm$ 1.3 & 66.1 $\pm$ 9.1 & 67.9 $\pm$ 12 & 51.0 $\pm$ 9.4 & 85.4 $\pm$ 0.6 & 82.4 $\pm$ 0.9 & 62.9 $\pm$ 2.6 & 66.1 $\pm$ 11 & 59.1 $\pm$ 12 & 48.4 $\pm$ 12 & 76.9 $\pm$ 1.0 & 73.5 $\pm$ 1.1 & 55.8 $\pm$ 3.3 & 59.3 $\pm$ 8.8 \\
            \midrule
            \multicolumn{19}{c}{\emph{Movies (OOD)}} \\
            \midrule
            \textbf{Baseline} & 70.0 $\pm$ 2.5 & 49.6 $\pm$ 1.2 & 58.1 $\pm$ 1.4 & - & - & - & 27.7 $\pm$ 2.2 & 18.5 $\pm$ 0.3 & 22.6 $\pm$ 0.8 & - & - & - & 28.2 $\pm$ 2.7 & 16.6 $\pm$ 0.4 & 22.8 $\pm$ 1.0 & - & - & - \\
            \textbf{ER + Att} & 74.6 $\pm$ 2.0 & 49.1 $\pm$ 0.7 & 58.1 $\pm$ 0.7 & - & - & - & 34.9 $\pm$ 2.9 & 18.4 $\pm$ 0.2 & 22.7 $\pm$ 0.5 & - & - & - & 35.3 $\pm$ 2.9 & 16.5 $\pm$ 0.2 & 23.1 $\pm$ 0.6 & - & - & - \\
            \textbf{ER + AttR} & 70.1 $\pm$ 1.7 & 50.2 $\pm$ 1.1 & 58.0 $\pm$ 1.0 & - & - & - & 27.5 $\pm$ 1.3 & 18.7 $\pm$ 0.3 & 22.5 $\pm$ 0.5 & - & - & - & 28.0 $\pm$ 1.5 & 16.8 $\pm$ 0.4 & 22.7 $\pm$ 0.6 & - & - & - \\
            \textbf{ER + IxG} & 69.9 $\pm$ 2.8 & 49.0 $\pm$ 1.3 & 59.5 $\pm$ 1.5 & - & - & - & 29.9 $\pm$ 2.6 & 18.4 $\pm$ 0.3 & 23.6 $\pm$ 1.0 & - & - & - & 30.7 $\pm$ 2.8 & 16.5 $\pm$ 0.4 & 24.1 $\pm$ 1.3 \\
            \textbf{ER-C + Att} & 68.8 $\pm$ 1.6 & 48.1 $\pm$ 0.6 & 56.0 $\pm$ 1.2 & - & - & - & 31.7 $\pm$ 1.6 & 18.2 $\pm$ 0.1 & 21.5 $\pm$ 0.6 & - & - & - & 32.6 $\pm$ 1.4 & 16.3 $\pm$ 0.2 & 21.6 $\pm$ 0.8 & - & - & - \\
            \textbf{ER-C + AttR} & 71.1 $\pm$ 2.0 & 62.8 $\pm$ 1.3 & 59.3 $\pm$ 1.1 & - & - & - & 31.3 $\pm$ 2.4 & 26.1 $\pm$ 0.8 & 23.1 $\pm$ 0.7 & - & - & - & 32.1 $\pm$ 2.7 & 26.6 $\pm$ 1.0 & 23.6 $\pm$ 0.9 & - & - & - \\
            \textbf{ER-C + IxG} & 56.7 $\pm$ 11 & 56.1 $\pm$ 3.6 & 61.6 $\pm$ 0.8 & - & - & - & 23.6 $\pm$ 5.6 & 20.6 $\pm$ 1.4 & 26.6 $\pm$ 0.8 & - & - & - & 22.7 $\pm$ 7.3 & 19.8 $\pm$ 2.1 & 28.2 $\pm$ 0.8 \\
            \midrule
            $\mathcal L_{\text{expl}}$ (A) & 61.0 $\pm$ 2.4 & 48.2 $\pm$ 1.5 & 48.0 $\pm$ 0.3 & - & - & - & 25.6 $\pm$ 0.6 & 18.0 $\pm$ 0.3 & 18.3 $\pm$ 0.1 & - & - & - & 27.4 $\pm$ 0.3 & 16.1 $\pm$ 0.4 & 17.0 $\pm$ 0.0 & - & - & - \\
            $\mathcal L_{\text{expl}}$ (R) & 59.4 $\pm$ 1.3 & 62.5 $\pm$ 0.4 & 55.2 $\pm$ 0.8 & - & - & - & 23.8 $\pm$ 0.3 & 25.8 $\pm$ 0.2 & 21.7 $\pm$ 0.4 & - & - & - & 22.7 $\pm$ 0.8 & 26.6 $\pm$ 0.2 & 21.2 $\pm$ 0.5 & - & - & - \\
            $\mathcal L_{\text{expl}}$ (IxG) & 51.8 $\pm$ 7.3 & 56.9 $\pm$ 4.0 & 59.8 $\pm$ 1.0 & - & - & - & 20.9 $\pm$ 2.6 & 21.2 $\pm$ 2.3 & 25.1 $\pm$ 0.6 & - & - & - & 19.6 $\pm$ 4.8 & 20.9 $\pm$ 3.6 & 26.8 $\pm$ 0.7 \\
            \midrule
            \multicolumn{19}{c}{\emph{Yelp-50 (OOD)}} \\
            \midrule
            \textbf{Baseline} & 51.9 $\pm$ 2.5 & 37.1 $\pm$ 1.1 & 51.5 $\pm$ 1.4 & 66.2 $\pm$ 2.1 & 59.5 $\pm$ 2.0 & 60.3 $\pm$ 1.4 & 28.5 $\pm$ 2.2 & 21.8 $\pm$ 0.3 & 31.1 $\pm$ 1.2 & 48.4 $\pm$ 3.9 & 41.7 $\pm$ 2.5 & 39.5 $\pm$ 2.1 & 23.2 $\pm$ 3.0 & 14.3 $\pm$ 0.6 & 26.3 $\pm$ 1.4 & 41.7 $\pm$ 2.7 & 36.6 $\pm$ 2.4 & 36.3 $\pm$ 1.7 \\
            \textbf{ER + Att} & 69.7 $\pm$ 2.0 & 36.4 $\pm$ 1.1 & 50.7 $\pm$ 1.2 & 67.4 $\pm$ 1.9 & 59.1 $\pm$ 0.9 & 59.3 $\pm$ 1.5 & 53.1 $\pm$ 1.8 & 21.7 $\pm$ 0.3 & 30.8 $\pm$ 1.2 & 50.5 $\pm$ 2.2 & 40.6 $\pm$ 1.5 & 38.0 $\pm$ 1.7 & 46.8 $\pm$ 2.2 & 14.0 $\pm$ 0.5 & 25.6 $\pm$ 1.4 & 43.9 $\pm$ 2.1 & 35.7 $\pm$ 1.4 & 35.2 $\pm$ 1.7 \\
            \textbf{ER + AttR} & 52.5 $\pm$ 3.0 & 37.6 $\pm$ 1.2 & 51.7 $\pm$ 0.8 & 67.1 $\pm$ 1.5 & 59.8 $\pm$ 1.2 & 60.5 $\pm$ 1.3 & 28.9 $\pm$ 2.7 & 22.0 $\pm$ 0.3 & 31.4 $\pm$ 1.2 & 49.8 $\pm$ 2.8 & 42.6 $\pm$ 1.7 & 40.5 $\pm$ 2.1 & 23.5 $\pm$ 3.9 & 14.4 $\pm$ 0.7 & 26.6 $\pm$ 1.3 & 42.8 $\pm$ 2.1 & 37.0 $\pm$ 1.3 & 36.7 $\pm$ 1.7 \\
            \textbf{ER + IxG} & 56.5 $\pm$ 3.3 & 37.3 $\pm$ 1.0 & 53.6 $\pm$ 2.4 & 66.6 $\pm$ 2.6 & 59.2 $\pm$ 1.5 & 59.9 $\pm$ 1.7 & 34.9 $\pm$ 4.5 & 21.9 $\pm$ 0.3 & 33.4 $\pm$ 2.8 & 50.1 $\pm$ 4.1 & 41.5 $\pm$ 2.4 & 39.3 $\pm$ 2.4 & 30.0 $\pm$ 5.1 & 14.3 $\pm$ 0.7 & 28.9 $\pm$ 3.1 & 42.9 $\pm$ 3.8 & 36.2 $\pm$ 2.1 & 35.7 $\pm$ 2.0 \\
            \textbf{ER-C + Att} & 76.0 $\pm$ 1.2 & 37.4 $\pm$ 1.2 & 51.3 $\pm$ 1.4 & 69.4 $\pm$ 2.7 & 58.6 $\pm$ 2.9 & 58.9 $\pm$ 2.5 & 56.4 $\pm$ 1.5 & 21.9 $\pm$ 0.3 & 31.1 $\pm$ 1.2 & 49.5 $\pm$ 2.7 & 39.2 $\pm$ 3.4 & 36.8 $\pm$ 2.8 & 50.9 $\pm$ 1.6 & 14.4 $\pm$ 0.7 & 26.2 $\pm$ 1.6 & 43.6 $\pm$ 2.8 & 35.4 $\pm$ 3.2 & 34.5 $\pm$ 2.9 \\
            \textbf{ER-C + AttR} & 67.2 $\pm$ 5.8 & 69.4 $\pm$ 1.5 & 58.8 $\pm$ 1.4 & 76.7 $\pm$ 1.6 & 65.5 $\pm$ 1.5 & 67.9 $\pm$ 1.6 & 49.4 $\pm$ 7.5 & 49.5 $\pm$ 1.5 & 37.8 $\pm$ 1.6 & 58.3 $\pm$ 1.8 & 47.7 $\pm$ 1.3 & 49.0 $\pm$ 1.3 & 43.4 $\pm$ 7.1 & 43.8 $\pm$ 1.4 & 34.1 $\pm$ 1.6 & 51.8 $\pm$ 1.8 & 42.3 $\pm$ 1.5 & 43.4 $\pm$ 1.5 \\
            \textbf{ER-C + IxG} & 60.8 $\pm$ 13 & 50.5 $\pm$ 6.3 & 69.8 $\pm$ 1.8 & 73.2 $\pm$ 2.6 & 59.2 $\pm$ 3.5 & 63.0 $\pm$ 5.1 & 41.3 $\pm$ 13 & 27.5 $\pm$ 3.3 & 52.3 $\pm$ 0.9 & 54.4 $\pm$ 2.8 & 39.4 $\pm$ 4.4 & 41.1 $\pm$ 5.7 & 35.3 $\pm$ 13 & 23.5 $\pm$ 5.5 & 47.4 $\pm$ 1.3 & 48.8 $\pm$ 2.8 & 35.7 $\pm$ 4.0 & 39.0 $\pm$ 5.5 \\
            \midrule
            $\mathcal L_{\text{expl}}$ (A) & 70.1 $\pm$ 2.1 & 38.5 $\pm$ 1.9 & 47.1 $\pm$ 0.7 & 62.5 $\pm$ 1.3 & 51.3 $\pm$ 1.3 & 49.4 $\pm$ 3.1 & 50.3 $\pm$ 2.5 & 21.8 $\pm$ 0.7 & 27.9 $\pm$ 0.4 & 41.2 $\pm$ 1.1 & 30.0 $\pm$ 1.0 & 27.8 $\pm$ 1.6 & 45.4 $\pm$ 1.7 & 14.6 $\pm$ 1.1 & 23.6 $\pm$ 0.4 & 36.3 $\pm$ 1.3 & 25.9 $\pm$ 1.0 & 23.1 $\pm$ 2.3 \\
            $\mathcal L_{\text{expl}}$ (R) & 62.6 $\pm$ 2.9 & 68.6 $\pm$ 1.0 & 56.8 $\pm$ 1.7 & 66.5 $\pm$ 1.3 & 59.0 $\pm$ 1.7 & 61.6 $\pm$ 0.2 & 39.3 $\pm$ 2.9 & 47.1 $\pm$ 1.8 & 35.2 $\pm$ 1.4 & 43.8 $\pm$ 1.7 & 37.1 $\pm$ 1.3 & 39.0 $\pm$ 0.4 & 35.7 $\pm$ 4.0 & 42.1 $\pm$ 2.0 & 31.9 $\pm$ 1.2 & 39.5 $\pm$ 2.1 & 33.7 $\pm$ 1.4 & 35.6 $\pm$ 0.8 \\
            $\mathcal L_{\text{expl}}$ (IxG) & 55.2 $\pm$ 12 & 58.3 $\pm$ 8.3 & 68.6 $\pm$ 0.7 & 72.2 $\pm$ 1.7 & 58.8 $\pm$ 1.6 & 60.7 $\pm$ 6.0 & 37.7 $\pm$ 10 & 33.4 $\pm$ 7.5 & 49.5 $\pm$ 0.2 & 53.6 $\pm$ 2.2 & 38.2 $\pm$ 2.0 & 41.1 $\pm$ 8.1 & 31.6 $\pm$ 11 & 32.2 $\pm$ 10 & 44.9 $\pm$ 1.0 & 47.1 $\pm$ 2.1 & 34.9 $\pm$ 2.5 & 38.3 $\pm$ 7.4 \\
            \bottomrule
        \end{tabular}
        \caption{\textbf{Attributions to Annotated Tokens}. AUC, Average Precision and Recall@k Scores ($\uparrow$) for SST-Dev (ID), Movies (OOD), and Yelp-50 (OOD) datasets, for the Baseline (Baseline), ER model guided with Attention (ER + Att), ER model guided with Attention-Rollout (ER + AttR), ER model guided with InputXGradient (ER + IxG) and the corresponding constrained counterparts (ER-C + X). Attributions are computed using Attention (\textsc{Att}), Attention + Rollout (\textsc{AttR}), InputXGradient (\textsc{IxG}), ALTI Aggregated (\textsc{ALTI}), and DecompX (with) Classifier (\textsc{DX} and \textsc{DX-C}) as the attribution techniques. For \textsc{Movies}, we report only the techniques that can be applied to longer sequences.}
        \label{tab:sa_plausibility_all}
        \end{minipage}
    }
\end{table*}

\end{document}